\documentclass[12pt]{article}

\usepackage{graphicx} 
\usepackage{amssymb}
\usepackage{amsfonts}
\usepackage{amsmath}
\usepackage{mathtools}
\usepackage{booktabs}
\usepackage{hyperref}
\usepackage{color}
\usepackage{algorithm}
\usepackage{algpseudocode}
\usepackage{longtable}
\usepackage[flushleft]{threeparttable}
\usepackage{cleveref}
\usepackage{subcaption}

\usepackage[font=scriptsize,labelfont=scriptsize]{caption}

\hypersetup{
    colorlinks,
    linkcolor={blue},
    citecolor={blue},
    urlcolor={blue}
}

\DeclarePairedDelimiter\norm{\lVert}{\rVert}%

\title{Cortex-Grounded Diffusion Models for Brain Image Generation}

\author{
Fabian Bongratz$^{\text{a,b}}$*, Yitong Li$^{\text{a,b}}$*, Sama Elbaroudy$^{\text{a}}$,\\
and Christian Wachinger$^{\text{a,b}}$ \\
\small $^{\text{a}}$Lab for AI in Medical Imaging, Technical University of Munich, Germany \\
\small $^{\text{b}}$Munich Center for Machine Learning, Germany \\
\small *Equal contribution \& correspondence: 
\texttt{\{fabi.bongratz, yi\_tong.li\}@tum.de}
}

\date{}

\begin{document}

\maketitle

\begin{abstract}

Synthetic neuroimaging data can mitigate critical limitations of real-world datasets, including the scarcity of rare phenotypes, domain shifts across scanners, and insufficient longitudinal coverage. However, existing generative models largely rely on weak conditioning signals, such as labels or text, which lack anatomical grounding and often produce biologically implausible outputs. To this end, we introduce \emph{Cor2Vox}, a cortex-grounded generative framework for brain magnetic resonance image (MRI) synthesis that ties image generation to continuous structural priors of the cerebral cortex. It leverages high-resolution cortical surfaces to guide a 3D shape-to-image Brownian bridge diffusion process, enabling topologically faithful synthesis and precise control over underlying anatomies. To support the generation of new, realistic brain shapes, we developed a large-scale statistical shape model of cortical morphology derived from over 33,000 UK Biobank scans. We validated the fidelity of Cor2Vox based on traditional image quality metrics, advanced cortical surface reconstruction, and whole-brain segmentation quality, outperforming many baseline methods. Across three applications, namely (i) anatomically consistent synthesis, (ii) simulation of progressive gray matter atrophy, and (iii) harmonization of in-house frontotemporal dementia scans with public datasets, Cor2Vox preserved fine-grained cortical morphology at the sub-voxel level, exhibiting remarkable robustness to variations in cortical geometry and disease phenotype without retraining.

 \end{abstract}

\section{Introduction}

\begin{figure}[hp]
    \centering
    \includegraphics[width=0.9\linewidth]{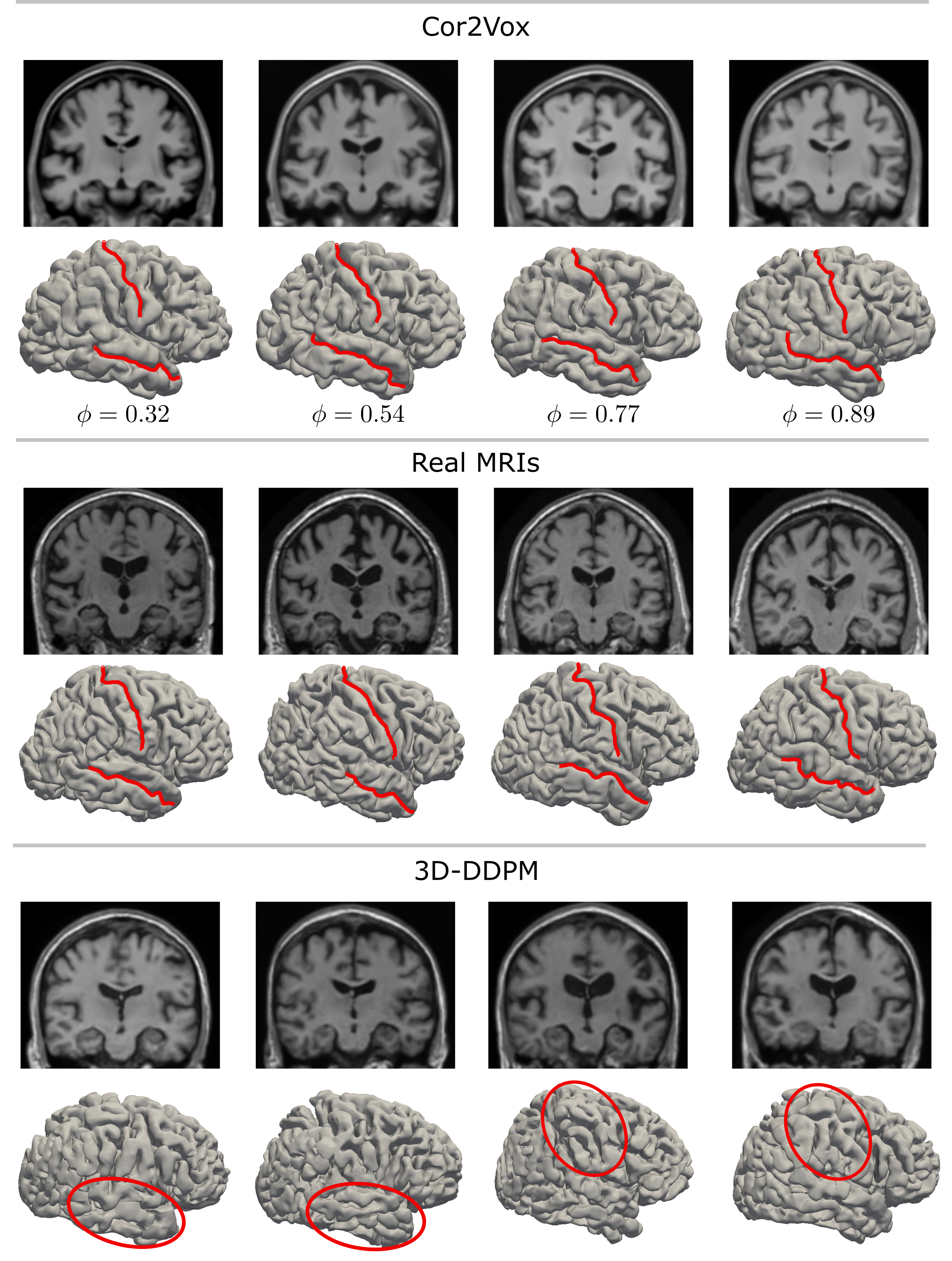}
    \caption{Synthetic MRIs generated by Cor2Vox (top) and 3D-DDPM (bottom), with real MRIs shown for reference (center). Cortical surfaces were reconstructed from the MRIs using Vox2Cortex-Flow. Red lines indicate the central sulcus and superior temporal sulcus, two prominent cortical grooves in human brain anatomy. While Cor2Vox reliably preserves these sulci, they are often absent or significantly distorted in surfaces reconstructed from MRIs generated by 3D-DDPM. Such anatomical irregularities are virtually impossible to detect in conventional 2D slice-based views and require 3D surface reconstructions for reliable assessment. $\phi\in(0,1)$ denotes the spherical interpolation factor in the Cor2Vox shape model.}
    \label{fig:realism}
\end{figure}

Synthetic data can overcome the persistent limitations of real-world medical datasets, such as the limited availability of rare phenotypes, the heterogeneity of imaging devices that introduces domain shifts, and the scarcity of longitudinal data required for disease progression modeling~\cite{vanBreugel2024synthetic}.
Despite these advantages, current approaches to medical image generation predominantly rely on implicitly learned associations between low-dimensional labels or short text and high-dimensional image observations~\cite{Sun2022HA-GAN,Khader2023denoising,Khosravi2024synthetic,Tudosiu2024morphology-preserving,Peng2024metadataconditioned,Wang2024self-improving}. This paradigm lacks explicit control over generated content, making it susceptible to undesired bias~\cite{Kocak2024biasinAI}, memorization of sensitive training data~\cite{Dar2025}, model collapse~\cite{Shumailov2024collapse}, and implausible hallucinations~\cite{Tivnan2024hallucination}. 
In neuroimaging, these challenges are amplified by the complexity of the cerebral cortex, which is inherently difficult to model due to its intricate folds and grooves. Structural inaccuracies and hallucinations often go unnoticed in conventional 2D visualizations and image-based metrics (see \Cref{fig:realism}).

Visual grounding~\cite{Guo2026grounding} and spatial conditioning~\cite{Zhang2023controlnet} have recently gained importance in computer vision by anchoring image generation to a structured, controllable signal as a potential solution to these challenges~\cite{Guo2026grounding,Zhang2023controlnet}. 
Initial work in medical imaging has explored controllable generation by conditioning on segmentation masks~\cite{Guo2025maisi,Deo2024vasculature}.
However, such approaches are not directly transferable to neuroimaging. 
In particular, voxel-level masks generated, for example, by a variational autoencoder (VAE)~\cite{Kingma2014}, typically do not guarantee anatomical plausibility and topological correctness. 
Moreover, brain-related phenotypes, such as those associated with aging or neurodegenerative disorders, manifest as subtle, sub-millimeter morphological changes in the brain structure~\cite{Bethlehem2022braincharts,Du2006corticalthinningalzheimerfrontotemporal}. Capturing these fine-grained variations requires high-resolution surface meshes rather than coarse voxel masks to ensure faithful representation of neuroanatomy~\cite{fischlFreeSurfer2012,cat12}.

In this study, we address these issues by introducing \emph{Cor2Vox}, a cortex-grounded generative framework for synthesizing brain magnetic resonance images (MRIs). By explicitly linking image generation to anatomically informed shape constraints of the cerebral cortex, we ensure that outputs remain biologically plausible and reliably reflect the variability observed in the population.
The image generation in Cor2Vox is rooted in the Brownian bridge diffusion process, originally proposed for 2D image-to-image translation~\cite{li_bbdm_2023}, which is extended to 3D spatially conditioned image synthesis. To enhance anatomical fidelity, we introduce auxiliary shape conditioning based on signed distance fields (SDFs) and structural masks during the reverse diffusion process, enabling precise guidance and alignment with cortical geometry. In contrast to conventional denoising diffusion probabilistic models (DDPMs)~\cite{ddpm}, which are based on unstructured Gaussian priors, the Brownian bridge diffusion in Cor2Vox maps directly from a dedicated cortex SDF to an output image, thereby eliminating the dependence on an unstructured latent space. 
As part of Cor2Vox, we present a large-scale statistical shape model of the human cerebral cortex, built upon deep learning-based mesh reconstructions of cortical boundaries with over 163,000 vertices per hemisphere. This shape model, derived from over 33,000 UK Biobank brain MRI scans~\cite{Sudlow2015ukb}, enables the synthesis of unseen, anatomically plausible cortical geometries by navigating the shape space that captures realistic brain anatomy in the population. Importantly, the generative capabilities of Cor2Vox surpass previous methods for simulating neuroanatomical change, which are typically constrained to morphing pre-existing scans~\cite{Karacali2006simulation,Sharma2010atrophy,Camara2006atrophy,CastellanoSmith2003simulation,Khanal2016biophysical,Rusak2022benchmark}, limiting their ability to generate unseen anatomies.

Our experiments showcase three practical applications of Cor2Vox. First, we utilize Cor2Vox to synthesize anatomically consistent brain MRIs that adhere to gyral folding patterns and reflect parcellation-based cortical thickness. Second, we conduct mesh-based continuous deformations to simulate progressive cortical atrophy at the sub-voxel level, which is important for studying neurodegenerative trajectories and benchmarking of cortical thickness estimation methods. Finally, we employ Cor2Vox to harmonize MR images across the ADNI database and in-house frontotemporal dementia~(FTD) cases. To this end, we leverage Cor2Vox's ability to disentangle brain morphology, which captures characteristic atrophy patterns in FTD, from its image appearance that is typically affected by scanner hardware and acquisition protocols.
Compared to our previous related work~\cite{cor2vox_ipmi}, we also provide a more extensive evaluation of Cor2Vox's image fidelity, using traditional image fidelity metrics, advanced cortical surface reconstruction based on Vox2Cortex-Flow~\cite{Bongratz2024v2cflow}, and automatic whole-brain segmentation quality assessment based on SynthSeg+~\cite{Billot2023synthseg}. 
Code to reproduce our results is available at \url{https://github.com/ai-med/Cor2Vox}.

\section{Materials and Methods}

\begin{figure}[hp]
    \centering
    \includegraphics[width=0.98\linewidth]{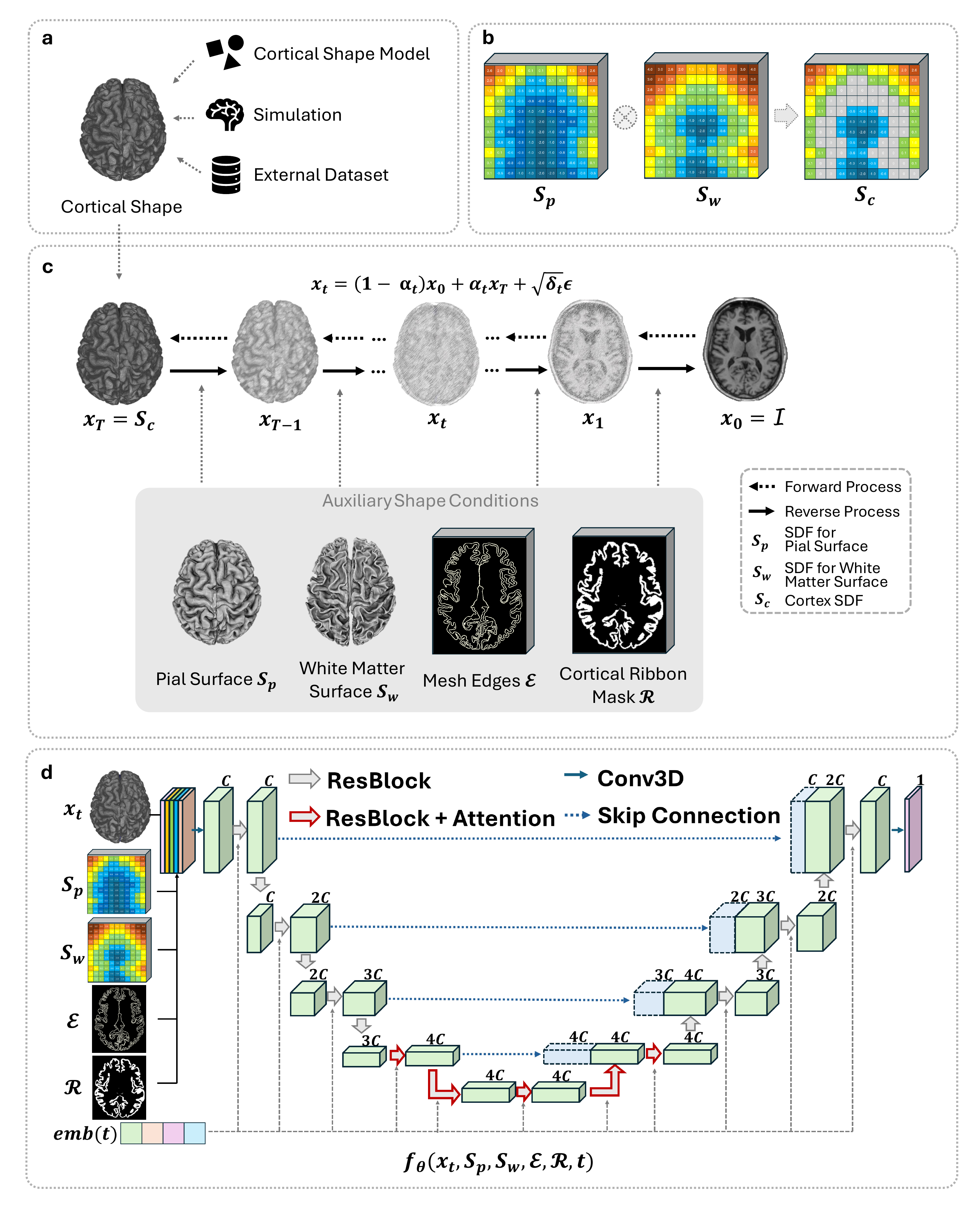}
    \caption{Cor2Vox overview. a, In dependence on the application, input cortical surfaces can be created based on a statistical shape model of the cerebral cortex, surface-based simulations, or existing shapes from external databases. b, The cortical surface meshes, i.e., pial ($\mathcal{S}_p$) and white matter ($\mathcal{S}_w$) surfaces, are converted into a joint cortex signed distance field (SDF), $\mathcal{S}_c$. c, Cor2Vox leverages a shape-to-image Brownian bridge diffusion process to learn a stochastic mapping $f_\theta$ from the spatial condition $\mathcal{S}_c$ to the output image $\mathcal{I}$. During the reverse diffusion process, auxiliary shape conditions are incorporated to improve geometric consistency with the input condition. d, A residual 3D UNet with convolutional and attention blocks is implemented for the prediction of the reverse diffusion process.}
    \label{fig:architecture}
\end{figure}

\subsection{Cortex-conditioned brain image generation} \label{sec:cortex-to-image-model}

\subsubsection{Overview of Cor2Vox}
An overview of Cor2Vox is presented in \Cref{fig:architecture}, illustrating the high-level workflow and model architecture. Conceptually, Cor2Vox synthesizes brain MR images conditioned on cortical surfaces that define the cerebral cortex shape a priori, serving as a spatial constraint for image generation. 
The first step (\Cref{fig:architecture}a)  defines the cortical shape, for which we propose three complementary strategies: 
(i) sampling from a population-based shape model, which allows high-resolution cortices to be generated and their variability to be modeled by traversing the cortical shape manifold in latent space,
(ii) applying simulated morphological alterations, such as localized cortical thinning, and
(iii) importing cortical surfaces from external datasets, enabling cross-dataset harmonization.

In the second step (\Cref{fig:architecture}b), the input cortical shape, represented as 3D surface meshes, is converted into a signed distance field (SDF) representation of the cortex, denoted $\mathcal{S}_c$.
This representation combines SDFs of the inner (white matter) boundary $\mathcal{S}_w$ and outer (pial) boundary $\mathcal{S}_p$. Based on this spatial condition, Cor2Vox generates an MRI via a Brownian bridge diffusion process (\Cref{fig:architecture}c). The forward process analytically maps between the image and the cortical condition, while the reverse diffusion is guided by auxiliary shape constraints, including $\mathcal{S}_p$, $\mathcal{S}_w$, a binarized edge map $\mathcal{E}$, and a cortical ribbon mask $\mathcal{R}$. 
\Cref{fig:architecture}d illustrates the denoising network architecture, which uses a 3D residual U-Net with convolutional and attention blocks, trained to predict $x_{t-1}$ from $x_t$ during reverse diffusion.

\subsubsection{Representing the cortex as a signed distance field} \label{sec:cortex-representation}

We propose to use an SDF-based representation of the cortex as a starting point for brain image generation, cf.\ \Cref{fig:architecture}.
For a closed surface $\mathcal{S}\subset\mathbb{R}^3$, an SDF is defined as 
\begin{equation} 
g: \Omega \subseteq \mathbb{R}^3 \to \mathbb{R}, 
\end{equation} 
where $g(x)$ denotes the signed orthogonal distance from a point $x \in \Omega$ to the surface $\mathcal{S}$. The sign convention typically assigns negative values to points inside $\mathcal{S}$ and positive values to points outside;
the zero-level set, i.e., $g(x)=0$, corresponds to the surface itself.
When sampled densely, SDFs align with the grid-like structure of 3D image data, a property that is essential for implementing the Brownian bridge-based diffusion process within our architecture.

However, representing the cortical ribbon with an SDF is not straightforward, as it is bounded by two closely spaced surfaces: the outer pial surface and the inner white matter surface. Each of these surfaces is traditionally provided as a separate mesh for each hemisphere. As a first preprocessing step, we merged the pial and white matter meshes into unified representations, denoted by $\mathcal{M}_p$ and $\mathcal{M}_w$, respectively.
From these meshes, we constructed a unified cortex SDF, $\mathcal{S}_c$, which serves as the primary source input for our shape-to-image framework.
To achieve this, we first converted the pial and white matter meshes into their respective dense SDFs, $\mathcal{S}_p$ and $\mathcal{S}_w$.
Next, we fused the two SDFs into a unified cortical representation, $\mathcal{S}_c$, following a region-specific logic:
i) Region outside the cortex: when both $\mathcal{S}_p$ and $\mathcal{S}_w$ are positive, we assign $\mathcal{S}_c$ the value of $\mathcal{S}_p$, which is the distance to the pial surface, i.e., the closer surface in this case.
ii) Cortex region: when $\mathcal{S}_p$ and $\mathcal{S}_w$ have opposite signs, the point is part of the cortex. Here, we set $\mathcal{S}_c$ to zero, creating a clear delineation of the cortical region. We also extracted this zone as a binary mask, $\mathcal{R}$.
iii) Region enclosed by the cortical ribbon: when both $\mathcal{S}_p$ and $\mathcal{S}_w$ are negative, $\mathcal{S}_c$ takes the value of $\mathcal{S}_w$, which is the distance to the white matter surface, i.e., the closer surface in this case.
The detailed procedure is also illustrated in Figure~S5.

Our goal is to synthesize a 3D MRI volume $\mathcal{I} \in \mathbb{R}^{H \times W \times D}$ that faithfully reflects the cortical geometry encoded by the SDF $\mathcal{S}_c$. Formally, this requires learning a mapping
\begin{equation} \label{eq:f-theta}
    f_\theta : \mathbb{R}^{H \times W \times D} \longrightarrow \mathbb{R}^{H \times W \times D},
\end{equation}
such that $\mathcal{I} = f_\theta(\mathcal{S}_c)$ parameterized by $\theta$. Inspired by recent advances in image-to-image translation~\cite{li_bbdm_2023,Lee2024ebdm}, we model $f_\theta$ using a neural network-based Brownian bridge diffusion process within our Cor2Vox framework, as illustrated in \Cref{fig:architecture}.

\subsubsection{Brownian bridge diffusion process}

In standard Denoising Diffusion Probabilistic Models (DDPMs)~\cite{ddpm}, incorporating spatial conditions is inherently challenging because generation typically begins from pure Gaussian noise, $\varepsilon \sim \mathcal{N}(0, \mathbf{I})$. In contrast, the Brownian Bridge Diffusion Model (BBDM)~\cite{li_bbdm_2023} establishes a direct mapping between structured source and target distributions. In our setting, the Brownian bridge connects the cortex SDF $\mathcal{S}_c$ to its corresponding MRI volume $\mathcal{I}$, enabling a direct, anatomically informed conditioning in the diffusion process. This formulation allows training directly on paired samples $(\boldsymbol{x}_T, \boldsymbol{x}_0) \sim q_{\text{data}}(\mathcal{S}_c, \mathcal{I})$, thereby eliminating the need for Gaussian noise as an initial state.\\

\noindent
\textbf{Forward diffusion process.} Beginning with an initial state $\boldsymbol{x}_0$ (MRI volume $\mathcal{I}$), the forward process progressively transforms it to the destination state $\boldsymbol{x}_T$ (shape prior $\mathcal{S}_c$) over a series of stochastic timesteps. This process is defined as follows:
\begin{equation}
    q(\boldsymbol{x}_t \mid \boldsymbol{x}_0, \boldsymbol{x}_T) = \mathcal{N}\big(\boldsymbol{x}_t; (1 - \alpha_t)\boldsymbol{x}_0 + \alpha_t \boldsymbol{x}_T, \delta_t \mathbf{I}\big), \quad \boldsymbol{x}_T \equiv \mathcal{S}_c ,
\end{equation}
where $\alpha_t = t/T$, $T$ is the total number of steps in the diffusion process, and $\delta_t = 2(\alpha_t - {\alpha_t}^2)$ is the variance term. In analogy to~\cite{li_bbdm_2023}, the transition probability between two consecutive steps in the Brownian bridge is given as:
\begin{align}
\label{eq:bbdm_forward}
    q_{\text{BB}}(\boldsymbol{x}_t \mid \boldsymbol{x}_{t-1}, \boldsymbol{x}_T) &= \mathcal{N}\bigg(\boldsymbol{x}_t; \frac{1 - \alpha_t}{1 - \alpha_{t-1}} \boldsymbol{x}_{t-1} + \left(\alpha_t - \frac{1 - \alpha_t}{1 - \alpha_{t-1}} \alpha_{t-1}\right) \boldsymbol{x}_T, \delta_{t \mid t-1} \mathbf{I}\bigg), \nonumber \\
    \delta_{t \mid t-1} &= \delta_t - \delta_{t-1} \frac{(1 - \alpha_t)^2}{(1 - \alpha_{t-1})^2}.
\end{align}

\begin{algorithm}[t]
\caption{Training Process of Cor2Vox}
\label{alg:training}
\begin{algorithmic}[1]
\Repeat
    \State Paired data $\boldsymbol{x}_0 \sim q(\mathcal{I}), \, \boldsymbol{x}_T \sim q(\mathcal{S}_c)$ and auxiliary conditions $\mathcal{C} = (\mathcal{S}_p, \mathcal{S}_w, \mathcal{E}, \mathcal{R}) \sim q(\mathcal{S}_p, \, \mathcal{S}_w, \mathcal{E}, \mathcal{R})$
    \State Timestep $t \sim \text{Uniform}(1, \dots, T)$
    \State Gaussian noise $\varepsilon \sim \mathcal{N}(0, \mathbf{I})$
    \State Forward diffusion $\boldsymbol{x}_t = (1 - \alpha_t)\boldsymbol{x}_0 + \alpha_t \boldsymbol{x}_T + \sqrt{\delta_t} \varepsilon$
    \State Take gradient descent step on
    $\nabla_\theta 
    \lVert \alpha_t (\boldsymbol{x}_T - \boldsymbol{x}_0) 
    + \sqrt{\delta_t} \boldsymbol{\varepsilon} - f_\theta(\boldsymbol{x}_t, \mathcal{C}, t) \rVert_1$
\Until{converged}
\end{algorithmic}
\end{algorithm}

\begin{algorithm}[t]
\caption{Sampling Process of Cor2Vox}
\begin{algorithmic}[1]
\State Sample conditional input $x_T = \mathcal{S}_c \sim q(\mathcal{S}_c)$, $\mathcal{C} = (\mathcal{S}_p, \mathcal{S}_w, \mathcal{E}, \mathcal{R}) \sim q(\mathcal{S}_p, \mathcal{S}_w, \mathcal{E}, \mathcal{R})$
\For{$t = T, \dots, 1$}
    \State \textbf{if} $t > 1$ \textbf{then} $\varepsilon \sim \mathcal{N}(0, \mathbf{I})$, \textbf{else} $\varepsilon = 0$
    \State $x_{t-1} = c_{xt} \boldsymbol{x}_t + c_{st} \mathcal{S}_c - c_{ft} f_\theta \big(\boldsymbol{x}_t, \mathcal{C}, t \big) + \sqrt{\tilde{\delta}_t} \varepsilon$
\EndFor
\State \Return $x_0$
\end{algorithmic}
\label{alg:sampling}
\end{algorithm}

\noindent
The variance term $\delta_t$ follows a bridge-like schedule. At the start of the diffusion process ($t = 0$), it is set to zero, $\delta_0 = 0$, ensuring the process begins from the MRI image, i.e., $\boldsymbol{x}_0 \equiv \mathcal{I}$. As diffusion progresses, $\delta_t$ gradually increases, reaching its maximum value $\delta_{\text{max}} = \delta_{T/2} = 1/2$ at the midpoint, introducing the highest level of stochasticity during the transition between image and shape. Beyond this point, the variance decreases symmetrically, returning to zero at the final timestep ($t = T$), where $\alpha_T = 1$, guaranteeing convergence to the target SDF $\mathcal{S}_c$. This schedule enables a smooth yet stochastic transformation from the image domain $\mathcal{I}$ to the shape domain $\mathcal{S}_c$, effectively learning a continuous path between two fixed states and establishing a direct link between anatomical shape and its visual representation. \\

\noindent
\textbf{Reverse diffusion process.} The reverse Brownian bridge process begins directly at the input condition $\boldsymbol{x}_T \equiv \mathcal{S}_c$ and ends at the MRI ($
\mathcal{I}$), essentially reversing the forward process by predicting $\boldsymbol{x}_{t-1}$ from $\boldsymbol{x}_t$.
However, we propose to incorporate additional structural representations of the cortex as auxiliary inputs to the denoising model. Specifically, these auxiliary representations include i) the pial surface SDF $\mathcal{S}_p$, ii) the white matter surface SDF $\mathcal{S}_w$, iii) a binary edge map $\mathcal{E}$ capturing pial and white matter boundaries, and iv) the cortical ribbon mask $\mathcal{R}$, cf. \Cref{sec:cortex-representation}. These conditions are jointly represented as $\mathcal{C} = (\mathcal{S}_p, \mathcal{S}_w, \mathcal{E}, \mathcal{R})$ and provided to the model at each timestep $t$ of the reverse process, which we found to improve the geometric consistency considerably, cf.\ \Cref{sec:ablation-results}.

Considering the additional guidance by the auxiliary conditions, the transition probability in our reverse process is given as:
\begin{equation}
\label{eq:bbdm_reverse}
    p_\theta(\boldsymbol{x}_{t-1} \mid \boldsymbol{x}_t, \mathcal{C}, \mathcal{S}_c) = \mathcal{N}\big(\boldsymbol{x}_{t-1}; \boldsymbol{\mu}_\theta(\boldsymbol{x}_t, \mathcal{C}, t), \tilde{\delta}_t \mathbf{I}\big),
\end{equation}
\noindent 
where $\boldsymbol{\mu}_\theta(\boldsymbol{x}_t, \mathcal{C}, t)$ represents the predicted mean value and $\tilde{\delta}_t$ denotes the noise variance at each timestep. In our implementation, we follow the reparameterization strategy used in DDPM~\cite{ddpm}, which trains a neural network $f_\theta(\cdot)$ to predict solely the noise rather than the mean $\boldsymbol{\mu}_\theta$. Precisely, we reformulate $\boldsymbol{\mu}_\theta$ as a linear combination of $\boldsymbol{x}_t$,  $\mathcal{S}_c$, and the estimated part $f_\theta$:
\begin{align}
\label{eq:bbdm_ep}
\boldsymbol{\mu}_\theta(\boldsymbol{x}_t, \mathcal{C}, \mathcal{S}_c, t) &= c_{xt} \boldsymbol{x}_t + c_{st} \mathcal{S}_c + c_{ft} f_\theta(\boldsymbol{x}_t, \mathcal{C}, t), \, \text{where} \\
    c_{xt} &= \frac{\delta_{t-1}}{\delta_t} \frac{1 - \alpha_t}{1 - \alpha_{t-1}} + \frac{\delta_{t|t-1}}{\delta_t} (1 - \alpha_{t-1}), \nonumber  \\
    c_{st} &= \alpha_{t-1} - \alpha_t \frac{1 - \alpha_t}{1 - \alpha_{t-1}} \frac{\delta_{t-1}}{\delta_t}, \nonumber \, \text{and}  \\
    c_{ft} &= (1 - \alpha_{t-1}) \frac{\delta_{t|t-1}}{\delta_t}. \nonumber
\end{align}
The parameters $c_{xt}$, $c_{st}$, and $c_{ft}$ are non-trainable since they are derived from $\alpha_t$, $\alpha_{t-1}$, $\theta_t$, and $\theta_{t-1}$.
The variance term $\tilde{\delta}_t$ does not need to be learned either; it can be derived analytically as $\tilde{\delta}_t = \frac{\delta_{t|t-1} \delta_{t-1}}{\delta_t}$.\\

\noindent
\textbf{Training.} The training process is designed to minimize the disparity between the joint distribution predicted by $f_\theta$ and the training data. This is accomplished by optimizing the Evidence Lower Bound (ELBO),
which provides a lower bound on the log-likelihood of the data and is defined as follows:
\begin{align}
\text{ELBO} &= - \mathbb{E}_q \big( \text{D}_{\text{KL}}(q_{\text{BB}}(\boldsymbol{x}_T \vert ~\boldsymbol{x}_0, \mathcal{S}_c) \parallel p(\boldsymbol{x}_T \vert ~\mathcal{C}, \mathcal{S}_c))  \nonumber \\
&\quad + \sum_{t=2}^T \text{D}_{\text{KL}}(q_{\text{BB}}(\boldsymbol{x}_{t-1} \vert ~\boldsymbol{x}_t, \boldsymbol{x}_0, \mathcal{S}_c) \parallel p_\theta(\boldsymbol{x}_{t-1} \vert ~\boldsymbol{x}_t, \mathcal{C}, \mathcal{S}_c)) \nonumber \\
&\quad - \log p_\theta(\boldsymbol{x}_0 \vert ~\boldsymbol{x}_1, \mathcal{C}, \mathcal{S}_c) \big).
\end{align}
By combining the ELBO with Eq.~(\ref{eq:bbdm_forward})(\ref{eq:bbdm_reverse})(\ref{eq:bbdm_ep}), our loss function is given by:
\begin{equation}
    \mathcal{L}_{\text{Cor2Vox}} =
    \mathbb{E}_{x_0, \mathcal{S}_c, \varepsilon \sim \mathcal{N}(0, \mathbf{I})} \Big[
    \Big\lVert \alpha_t (\mathcal{S}_c - \boldsymbol{x}_0) 
    + \sqrt{\delta_t} \boldsymbol{\varepsilon} - f_\theta(\boldsymbol{x}_t, \mathcal{C}, t) \Big\rVert_1
    \Big].
\end{equation}
By minimizing this loss at each timestep, the model learns to reverse the forward diffusion process, i.e., to generate brain images based on a cortical shape condition. 
For an algorithmic description of the training process, see \Cref{alg:training}. 
\\

\noindent
\textbf{Sampling.} Based on the reverse diffusion formulation, Cor2Vox generates new images by adhering to the following iteration:
\begin{equation}
    \boldsymbol{x}_{t-1} = c_{xt} \boldsymbol{x}_t + c_{st} \mathcal{S}_c - c_{ft} f_\theta \big(\boldsymbol{x}_t, \mathcal{C}, t \big) + \sqrt{\tilde{\delta}_t} \varepsilon,
\end{equation}
\noindent where $\varepsilon \sim \mathcal{N}(0, \mathbf{I})$ when $t > 1$, otherwise $\varepsilon = 0$. We accelerate the sampling process by employing the DDIM (Denoising Diffusion Implicit Models)~\cite{ddim} strategy. This approach utilizes a non-Markovian process, enabling faster and more direct sampling while preserving the same marginal distributions as the original Markovian inference. For an algorithmic description of the sampling process, see \Cref{alg:sampling}. 

\subsubsection{Model architecture and training parameters}
To parameterize $f_\theta$, we utilize a 3D UNet architecture, based on the ADM model~\cite{DMbeatsGAN} and adapted for 3D volumetric data. 
We set the channel sizes for the four residual stages to $[64, 128, 192, 256]$. Our model further leverages global attention with 4 heads and 64 channels at a downsampling factor of 8. To integrate the timestep information, we use adaptive group normalization within each residual block.
In our experiments, we trained for 400 epochs on a single NVIDIA H100 94GB GPU using the Adam optimizer~\cite{2015-kingma-adam}. We set the initial learning rate to $1 \times 10^{-4}$, which is reduced by a factor of $0.5$ upon reaching a plateau. We used a batch size of $2$, and an exponential moving average (EMA) rate of $0.995$. For the diffusion process, we used 1,000 timesteps for training. For inference, we adopt the DDIM sampling strategy~\cite{ddim} with only $10$ timesteps, which represents an effective balance between image quality and computational efficiency.

\subsubsection{Cortical shape model} \label{sec:shape-model}
To promote anatomical plausibility in generated brain images while capturing realistic population-level variability, we propose to anchor the generative process to an explicit statistical shape model of the cerebral cortex. Samples drawn from this shape model serve directly as inputs to our cortex-to-image module described in \Cref{sec:cortex-to-image-model}, ensuring that generated images remain consistent with biologically informed structural priors. To this end, we developed an approach based on principal component analysis (PCA), which we will describe in the following.\\

\noindent
\textbf{Creating a shape model for the cerebral cortex.}
PCA-based shape models are well-established for anatomical structures such as bones and organs~\cite{Bhalodia2024deepssm,Adams2022spatiotemporal,Bastian2023s3m}. However, these structures are commonly represented by comparably small meshes or sets of landmarks, rarely exceeding 10,000 points. In contrast, modeling the cerebral cortex requires over 160,000 vertices per brain hemisphere, as provided by V2C-Flow~\cite{Bongratz2024v2cflow}.
Moreover, the ribbon-like structure introduces additional challenges that prevents the direct application of existing shape modeling approaches that were usually designed for single-surface structures.

Starting from pre-registered T1-weighted MRI scans in MNI152 standard space, V2C-Flow produces white matter and pial cortical surfaces with inherent vertex-wise correspondence to the FsAverage template (163,842 vertices)~\cite{Fischl1999fsaverage}. We compute midthickness surfaces $\mathcal{M}^{\text{mid}} \in \mathbb{R}^{163{,}842 \times 3}$, not provided by default, by averaging the coordinates of corresponding pial and white matter vertices, while preserving the original mesh connectivity. Additionally, vertex-wise cortical thickness $\mathcal{T} \in \mathbb{R}^{163{,}842}$ is estimated via bidirectional vertex-to-face distance measurements between the two surfaces. Together, the midthickness geometry and cortical thickness form a compact representation of the cortical ribbon, which integrates seamlessly into a PCA-based shape model. We fit separate PCA models for each hemisphere, retaining 8,192 latent components per side. This configuration explains over 95\% of population-level variability in the UKB dataset and achieves an average point-wise mesh reconstruction error (L2) below 1 mm on the held-out validation set.\\

\noindent
\textbf{Traversing the manifold of cortical shapes.}
To sample new cortical geometries, we propose a spherical interpolation (slerp)~\cite{shoemake1985animating}-based approach in PCA latent space. Thereby, we ensure that the sampled left and right hemispheres fit together, and we stay on the manifold of realistic brain anatomies. See the provided video in the supplementary material for a visualization of this approach.

First, we embed two randomly chosen cortices, given as tuples $(\mathcal{M}^{\text{mid}}_1, \mathcal{T}_1)$, $(\mathcal{M}^{\text{mid}}_2, \mathcal{T}_2)$, via PCA:
\begin{equation}
    \mathbf{e}_1, \, \mathbf{e}_2 = \text{PCA}((\mathcal{M}^{\text{mid}}_1, \mathcal{T}_1)), \, \text{PCA}((\mathcal{M}^{\text{mid}}_2, \mathcal{T}_2)); \quad \mathbf{e}_1, \mathbf{e}_2 \in \mathbb{R}^{8192}.
\end{equation}
Based on the normalized embeddings $\hat{\mathbf{e}}_1=\tfrac{\mathbf{e}_1}{\norm{\mathbf{e}_1}}$, $\hat{\mathbf{e}}_2=\tfrac{\mathbf{e}_2}{\norm{\mathbf{e}_2}}$, an arbitrarily chosen interpolation factor $\phi\in(0,1)$, angle $\beta=\arccos(\hat{\mathbf{e}}_1^T\hat{\mathbf{e}}_2)$, and radius $r=\norm{\mathbf{e}_1}$, we obtain a new embedding
\begin{equation}
    \mathbf{e}_\phi = r \, \text{slerp}(\mathbf{e}_1, \mathbf{e}_2) = r\left( \frac{\sin((1 - \phi)\beta)}{\sin(\beta)} \mathbf{e}_1 + \frac{\sin(\phi \beta)}{\sin(\beta)} \mathbf{e}_2 \right),
\end{equation}
and, finally, the corresponding cortex representation from the inverse PCA:
\begin{equation}
    (\mathcal{M}^{\text{mid}}_\phi, \mathcal{T}_\phi) = \text{PCA}^{-1}(\mathbf{e}_\phi).
\end{equation}
Compared to a standard linear interpolation, the spherical interpolation avoids the bias toward the population mean, which would result in overly smooth and therefore implausible cortical surfaces (see Figure~S3).

\subsection{Simulating cortical atrophy} \label{sec:cortical-thinning}

Simulating cortical atrophy, i.e., the progressive thinning and loss of cortical tissue over time, in MRIs is challenging since the longitudinal alterations in the cerebral cortex typically lie considerably below the image resolution of 1 mm~\cite{Shaw2016thinning,Kuperberg2003schizophrenia,Singh2006corticalthinningalzheimers}. Yet, the surface-based approach and the deliberate combination of white matter and pial surfaces in Cor2Vox enable fine-grained continuous simulations. Specifically, we deform pial surfaces towards white matter surfaces by moving pial vertices iteratively in the direction of surface normals.
The deformed surfaces can be used directly as input for the cortex-to-image module in Cor2Vox.

\subsection{Implementation of baseline methods} \label{sec:baseline-implementation}
For best comparison to Cor2Vox, we implemented the baseline methods used to benchmark the performance of our model in a consistent manner. Specifically, we extended Pix2Pix~\cite{pix2pix2017} and BBDM~\cite{li_bbdm_2023}, which were originally developed for 2D images, to 3D volumetric data. For BBDM, we further applied its diffusion process directly in image space instead of a latent space, which aligns with the implementation of Med-DDPM~\cite{Dorjsembe2024medddpm} and Cor2Vox.
Med-DDPM~\cite{Dorjsembe2024medddpm} is a dedicated 3D diffusion model for brain MRI generation based on conditional segmentation masks. 
In our implementation of Med-DDPM, we switched from slice-wise to volumetric intensity scaling and changed the denoising target from noise to the denoised image, which considerably enhanced its performance.   
For all baseline models, we trained two versions, based on the mesh edge maps ($\mathcal{E}$) and cortical ribbon masks ($\mathcal{R}$). These representations closely resemble the conditions employed in their original works. 
In addition, to isolate the effect of the Brownian bridge process, we developed a variant of our model, namely Cor2Vox{\scriptsize/DDPM}. In this variant, we replaced the Brownian bridge process with a standard denoising diffusion process, while keeping all other components, including the auxiliary shape conditions. This allows us to directly compare the efficacy of the two diffusion processes for our application.

\subsection{Evaluation metrics}

To thoroughly evaluate the performance of our model, we assessed the quality of the generated MRIs using a combination of traditional image fidelity metrics, cortical surface-based geometric consistency, and automated image quality assessment. \\

\noindent\textbf{Image fidelity.}
In our evaluation, we used two standard metrics for traditional image comparisons, Peak Signal-to-Noise Ratio (PSNR) and Structural Similarity Index Measure (SSIM). For unpaired data, where a direct ground-truth comparison is not possible (relevant only for experiments in \Cref{sec:ftd-results}), we computed a Multi-Reference SSIM~(MR-SSIM) score. This score is calculated by averaging the SSIM between a generated image and ten randomly selected images from the reference dataset.\\

\noindent
\textbf{Surface accuracy.}
To evaluate the consistency of generated MRIs with the cortex condition, we use deep learning-based cortical surface reconstruction with V2C-Flow~\cite{Bongratz2024v2cflow}.
Specifically, we compute the Average Symmetric Surface Distance (ASSD) between the input surfaces $\mathcal{S}$ and surfaces extracted by V2C-Flow from the synthetic images $\hat{\mathcal{S}}$:

\begin{equation}
  \mathrm{ASSD}(\hat{\mathcal{S}}, \mathcal{S}) = 
  \frac{\sum_{p \in \hat{\mathcal{P}}} d(p,\mathcal{S}) + \sum_{p \in \mathcal{P}} d(p,\hat{\mathcal{S}})}
  {|\hat{\mathcal{P}}| + |\mathcal{P}|},
\end{equation}
where $\hat{\mathcal{P}}\subset\hat{\mathcal{S}}$ and $\mathcal{P}\subset{S}$ are 100,000 surface points, respectively. 
The point-face distance $d(p, \mathcal{M})$ measures the orthogonal distance from a point $p\in \mathbb{R}^3$ to its closest point on the surface mesh $\mathcal{M}$ (lower is better). \\

\noindent
\textbf{Automatic image quality assessment.}
To validate the practical utility of our generated MRIs, we also assess their quality for downstream image processing. We use the automatic quality control built into the SynthSeg+ framework~\cite{billot2023synthseg+}, which calculates the quality of cortical and subcortical segmentations derived from the generated brain images in the form of an estimated Dice coefficient between 0 and 1 (higher is better).

\subsection{Computational software and hardware}

Our software is based on Python (v3.10.14) and PyTorch (v2.1.0). During the preparation of the manuscript, we used several other Python libraries to support data analysis, including pandas (v2.2.1), scipy (v1.10.0), torchvision (v0.16.0), scikit-learn (v1.0.2), nibabel (v5.2.1), monai (v1.3.0), torchio (v0.19.6), pytorch3d (v0.6.1), and trimesh (v4.6.5). Image processing and training of V2C-Flow involved FreeSurfer (v7.2). 
Training Cor2Vox on a single NVIDIA H100 GPU on a shared computing cluster had an average runtime of 2.16 seconds per iteration, whereas the synthesis during inference took around a minute per instance. For fitting the PCA-based shape model, we used a high-performance computing infrastructure with 2TB of main memory, which took approximately two days per PCA.

\subsection{Data and preprocessing}
In this study, we used data from the Alzheimer's Disease Neuroimaging Initiative~(ADNI)~\cite{Petersen2010adni}.
Specifically, we used T1-weighted brain magnetic resonance images~(MRIs) from cognitively normal subjects, subjects diagnosed with mild cognitive impairment~(MCI), and Alzheimer's disease~(AD). All scans were registered to MNI152 standard space via affine registration using NiftyReg (v1.5.69). Cortical surfaces were then extracted with a pre-trained Vox2Cortex-Flow (V2C-Flow)~\cite{Bongratz2024v2cflow} model, using identical train/validation/test splits, comprising 1,155/169/323 samples, respectively. We used only baseline scans from ADNI to avoid subject bias. 

For the cortical shape model, we used data from the UK Biobank~(UKB) Imaging Study~\cite{littlejohns2020uk}. 
Initially, we used 1,000 scans for fine-tuning the V2C-Flow model based on previously extracted FreeSurfer~\cite{fischlFreeSurfer2012} surfaces. The remaining 
scans were then processed with this fine-tuned V2C-Flow model, resulting in 38,432 white matter and pial surfaces for each brain hemisphere. After excluding outliers based on an initially fitted PCA model and a per-component threshold of 1,000 and keeping 10\% of samples for validation, 33,403 UKB scans remained for fitting our cortical PCA model.

Lastly, we used 10 in-house clinical scans obtained by Klinikum Rechts der Isar (TUM Klinikum, Munich, Germany). These scans are from 10 subjects diagnosed with behavioral-variant frontotemporal dementia (bvFTD). The diagnosis was established using clinical criteria, FDG-PET metabolism information, and cerebrospinal fluid biomarkers.

We created signed distance fields (SDFs) for the white matter and pial surfaces from the V2C-Flow meshes using a KDTree~\cite{Maneewongvatana2002kdtree} implemented in Scipy~(v1.10.0). Both brain hemispheres were combined. Internally, all models use an internal resolution of $128^3$ voxels. For evaluation, we rescaled all outputs back to MNI space, 1 mm isotropic resolution. For the computation of PSNR and SSIM metrics, we used SynthStrip~\cite{Hoopes2022synthstrip} to skull-strip both the generated images and the original images (\texttt{orig.mgz} files from FreeSurfer) to prevent scanner noise in the background of the original scans from confounding the evaluation. Despite this, all models were trained to produce the entire MRI, including the skull and background, which is reflected in our visualizations.

\section{Results}

\begin{figure}[htp]
    \centering
    \includegraphics[width=1\linewidth]{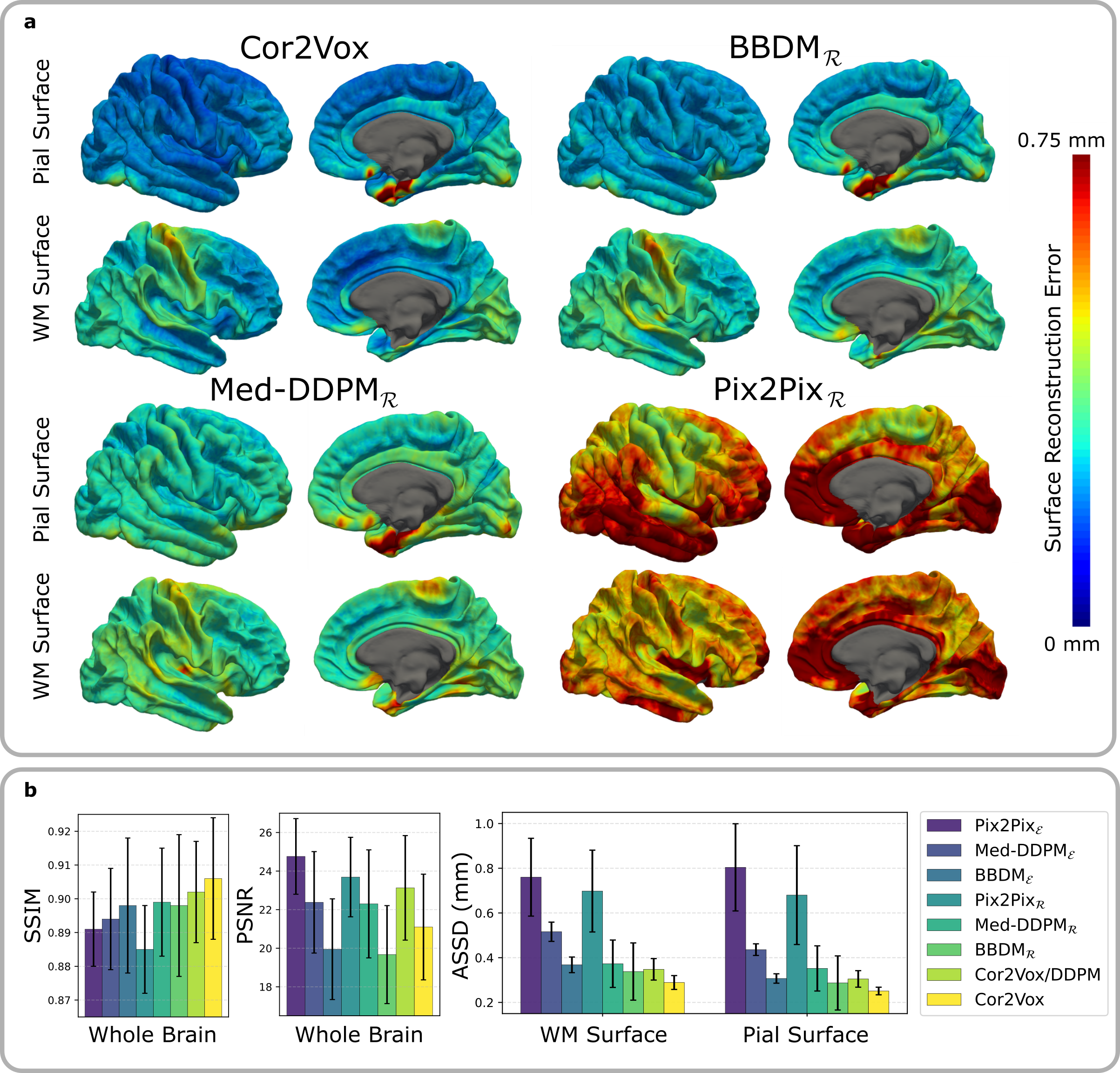}
    \caption{Image fidelity comparison of implemented methods for 3D brain MRI synthesis based on conditional cortical surfaces from the ADNI test set ($n=323$). Pix2Pix and BBDM were adapted for 3D generation. $\mathcal{R}$ and $\mathcal{E}$ indicate cortical ribbon mask and edge map inputs, respectively. a, Average vertex-wise reconstruction errors between input conditions and white matter (WM) and pial cortical surfaces reconstructed from the synthetic MRIs (lower is better). Visualizations are based on the FsAverage template. b, Structural similarity index measure (SSIM$\uparrow$) and peak signal-to-noise ratio (PSNR$\uparrow$) for whole-brain image fidelity and average symmetric surface distance (ASSD$\downarrow$, mm) for geometric consistency of right hemisphere WM and pial cortical surfaces (see Table~S1 for left hemisphere values). Bar plots show the mean and standard deviation (error bar) across all samples in the test set. 
    }
    \label{fig:generation_quality}
\end{figure}

\subsection{Image fidelity comparison to baseline methods} \label{sec:comparison}

\begin{figure}[hp]
    \centering
    \includegraphics[width=\linewidth]{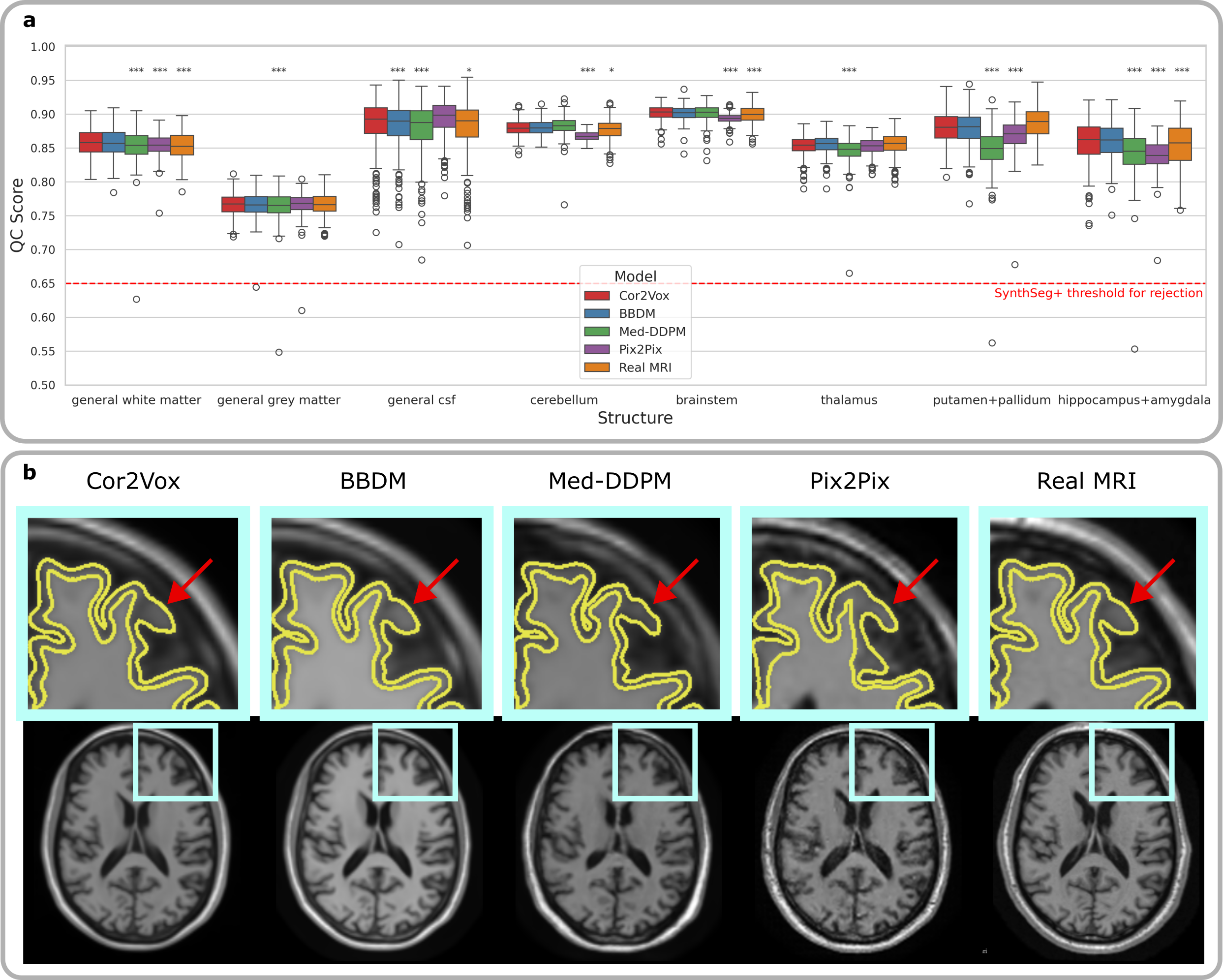}
    \caption{Brain segmentation quality and qualitative synthesis results. a, SynthSeg+-based automatic quality control (QC) scores of segmented brain structures in the synthetic MRIs. Asterisks indicate significant improvement of Cor2Vox over other methods as determined by a Wilcoxon signed rank test and Benjamini-Hochberg correction; ***: $p<0.001$, **: $p<0.01$, *: $p<0.05$. The horizontal red line indicates the recommended quality threshold for downstream processing of 0.65~\cite{billot2023synthseg+}. b, Synthetic MRIs from different methods and the real counterpart, visualized together with reconstructed cortical contours. All MRIs broadly show the same cortical geometry, with differences marked by red arrows.}
    \label{fig:comparison-baselines}
\end{figure}

\Cref{fig:generation_quality} reports the image fidelity of Cor2Vox in comparison to previous related baseline methods based on our ADNI test set (see also Table~S1). Specifically, we assessed the methods' abilities to preserve visual quality across the whole brain and maintain fine-grained geometric consistency with the cortical input conditions. We computed traditional quantitative metrics, including the structural similarity index measure (SSIM) and peak signal-to-noise ratio (PSNR), between the synthesized and ground-truth MRIs. Additionally, we reconstructed white matter (WM) and pial cortical surfaces from the synthetic MRIs using Vox2Cortex-Flow, which enabled sub-voxel evaluation of surface fidelity.

WM surfaces in Cor2Vox were, on average, 0.04\,mm more accurate than those from the second-best method, a 3D-extended version of the Brownian bridge diffusion model (BBDM)~\cite{li_bbdm_2023}, as measured by the average symmetric surface distance (ASSD, \Cref{fig:generation_quality}b). This represents an approximate 10\% improvement in surface accuracy, which was consistently observed across the entire cortex (see \Cref{fig:generation_quality}a). Similarly, the average reconstruction error in pial surfaces was 0.03\,mm lower in Cor2Vox compared to the BBDM. The improvements for both the WM and pial surfaces in terms of ASSD were significant, as indicated by a paired Wilcoxon signed-rank test ($p<10^{-7}$).   
A variant of our model, Cor2Vox{\scriptsize /DDPM}, which employs a DDPM-like diffusion process, outperformed Med-DDPM~\cite{Dorjsembe2024medddpm}, a dedicated model for spatially conditioned brain MRI generation, even after deliberate optimization of this method; see \Cref{sec:baseline-implementation} for implementation details. Yet, DDPM-based models could not keep up with the best Brownian bridge diffusion-based approaches (Cor2Vox and BBDM). Moreover, a 3D-extended version of Pix2Pix~\cite{pix2pix2017}, which leverages generative adversarial networks (GANs)~\cite{goodfellow2014gan}, was not competitive with the diffusion-based models in terms of geometric consistency. 
Overall, cortical ribbon masks ($\mathcal{R}$) consistently outperformed binarized edge maps ($\mathcal{E}$) as input conditions in our experiments, both of which are commonly used representations for spatial constraints in the original works presenting Pix2Pix, Med-DDPM, and BBDM. 

In terms of traditional whole-brain image fidelity metrics (\Cref{fig:generation_quality}b), Cor2Vox achieved the highest SSIM scores across all methods, while Pix2Pix yielded the best PSNR performance. However, prior studies have highlighted limitations of these conventional image metrics in the context of brain MRI synthesis~\cite{Wu2024evaluating,jafrasteh2025wasabi}, and have advocated for segmentation-based evaluation using SynthSeg~\cite{Billot2023synthseg,billot2023synthseg+}. Accordingly, we adopted SynthSeg+~\cite{billot2023synthseg+} for a more task-relevant assessment of image quality. The results, presented in \Cref{fig:comparison-baselines}a, demonstrate the superiority of Cor2Vox across many brain structures, with every individual sample surpassing the recommended quality threshold of 0.65 across all segmented brain structures. Finally, qualitative inspection of the synthesized MRIs, visualized in \Cref{fig:comparison-baselines}b, further shows the high cortical fidelity of Cor2Vox, with the reconstructed contours closely matching those of the original scan.

\begin{figure}[hp]
    \centering
    \includegraphics[width=0.9\linewidth]{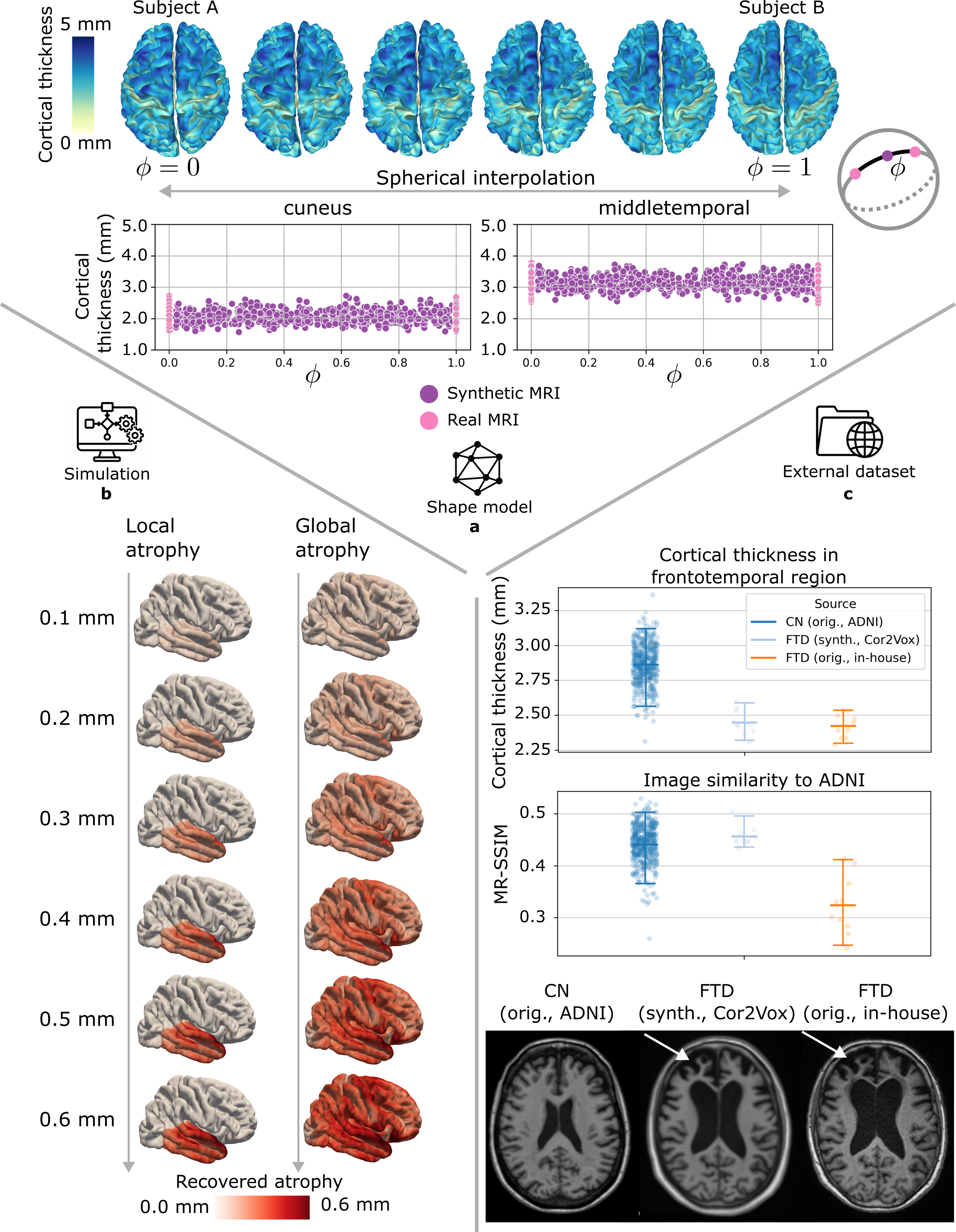}
    \caption{
    Results from three real-world applications of Cor2Vox.
    a, Surface plots visualize the anatomically consistent sampling of cortices by spherical interpolation in shape space for an arbitrary pair of UKB subjects. Scatter plots show cortical thickness in the cuneus and middle temporal regions of synthetic and real data based on randomly sampled subject pairs.
    b, Simulation of progressive cortical atrophy; surface plots show average recovered atrophy from 124 synthetic MRIs (1~mm isotropic resolution), in dependence on the respective introduced changes.
    c, Augmentation of the ADNI dataset with synthetic data from Cor2Vox, based on in-house cortical shapes from frontotemporal dementia (FTD) cases. A higher multi-reference (MR)-SSIM indicates a more similar image appearance to that of ADNI. Cor2Vox accurately preserved the reduced cortical thickness in the frontotemporal region, a hallmark of FTD~\cite{Du2006corticalthinningalzheimerfrontotemporal}, compared to ADNI normal controls (CN).
    }
    \label{fig:applications}
\end{figure}

\subsection{Anatomically consistent brain generation} \label{sec:plausibility-results}

While generative models can generate new brain images by sampling from abstract latent spaces, ensuring biological plausibility requires a population-based shape model. However, capturing cortical variability across individuals remains a major challenge due to the cortex’s intricate folding patterns, regional heterogeneity, and substantial inter-subject anatomical differences. To date, no comprehensive model of this variability exists. Leveraging cortical surface meshes from over 33,000 UK Biobank brain MRI scans, we constructed a detailed statistical model of cortical anatomy (see \Cref{sec:shape-model} for details). By traversing the latent space of this model via spherical interpolation, we ensure that sampled shapes lie on the manifold of anatomically realistic cortical geometry (cf.\ Figure~S3).

The sequence of cortical surfaces in \Cref{fig:applications}a illustrates this process, transitioning smoothly between two arbitrary UKB subjects (see also supplementary video). Importantly, the interpolation preserves both gyrification and cortical thickness, maintaining characteristic cortical grooves, such as the central sulcus, throughout. In contrast, as shown in \Cref{fig:realism} and Figure~S1, 3D-DDPMs failed to maintain such anatomical realism, often producing non-identifiable and discontinuous gyral patterns that hinder subsequent gyral-based parcellation, such as those based on the Desikan-Killiany (DK) atlas~\cite{desikan2006}. These parcellations are critical for numerous downstream neuroimaging applications. To assess anatomical plausibility, we applied the DK atlas to randomly generated, interpolation-based synthetic cortical shapes and compared regional cortical thickness against real counterparts. The results, summarized in the scatter plots in \Cref{fig:applications}a and Figure~S2, demonstrate that Cor2Vox consistently preserves anatomically realistic cortical thickness across all 34 DK regions.

Next, we compared the geometric consistency between sampled and real cortical shapes with their corresponding reconstructions from the synthesized MRIs. 
The average symmetric surface distances (ASSD) between sampled cortical surfaces ($n=323$) and corresponding reconstructions from the synthetic MRIs were 0.246~mm (SD: 0.019~mm) for the left hemisphere and 0.239~mm (SD: 0.016~mm) for the right hemisphere.
Conversely, ASSDs for real cortical surfaces from UKB were 0.255~mm (SD: 0.013~mm) and 0.258~mm (SD: 0.012~mm) per hemisphere. 
These results indicate that new, shape model-derived cortical surfaces integrate seamlessly into Cor2Vox’s image generation process, enabling the realistic synthesis of brains across the broad spectrum of cortical variability.

\subsection{Sub-voxel simulation of cortical atrophy} \label{sec:atrophy-results}

Controlled simulation of sub-voxel cortical atrophy is important for benchmarking methods for cortical surface reconstruction and cortical thickness estimation.
To evaluate the sensitivity of Cor2Vox to simulated alterations in cortical morphology, we mimicked progressive cortical thinning ranging from 0.1 to 0.6~mm (image resolution: 1~mm isotropic). As detailed in \Cref{sec:cortical-thinning}, atrophy was simulated once globally across the entire cortex and also locally in the temporal lobe. We used the cognitively normal control group from our ADNI test set as a basis for this experiment ($n=124$). From the synthesized scans, we estimated cortical thickness again with Vox2Cortex-Flow, comparing it to the original (baseline) cortical thickness for each subject. 
As shown in \Cref{fig:applications}b, the difference in cortical thickness estimated from the synthetic MRIs closely reflected the magnitude of the imposed changes. Specifically, we obtained a mean absolute error in recovered relative cortical thickness changes of 0.14~mm (SD: 0.08~mm) across all simulations of localized atrophy and, similarly, 0.14~mm (SD: 0.10~mm) for global gray matter atrophy.

\subsection{Shape-based image harmonization} \label{sec:ftd-results}

Neuroimaging datasets often focus on a single disease, such as Alzheimer's disease in ADNI. For studying clinically relevant questions, such as the differential diagnosis of dementing disorders, a combination of multiple datasets is needed; yet, this introduces unwanted dataset biases~\cite{wachinger2021detect}. 
Cor2Vox addresses this harmonization challenge by generating consistent MRIs from external cortical shapes. 
Importantly, scans from the target disease do not need to be included in the diffusion model's training set. 
We demonstrate this capability by generating MRI scans of patients with frontotemporal dementia~(FTD) from cortical shapes obtained from our in-house cohort using the ADNI-trained Cor2Vox. 

As shown in \Cref{fig:applications}c, the resulting synthetic scans integrate seamlessly into ADNI, exhibiting significantly higher mean Structural Similarity Index Measure (SSIM) to ten randomly chosen (multi-reference)
ADNI scans compared to the original in-house scans. Crucially, these synthesized MRIs retained the characteristic pathological features of FTD, exhibiting significantly lower cortical thickness in the frontotemporal region compared to cognitively normal (CN) subjects, thereby mirroring the characteristic morphology of this type of brain disease~\cite{Du2006corticalthinningalzheimerfrontotemporal}.

\begin{table}[hp]
    \setlength{\tabcolsep}{5pt}
    \renewcommand\bfdefault{b}
    \centering
    \caption{Influence of different source domains (Source) and auxiliary shape conditions (Aux.~Condition) provided to Cor2Vox during the Brownian bridge process. We report the mean\textpm SD on the ADNI validation set ($n=169$). Geometric accuracy is measured in terms of average symmetric surface distance (ASSD) in mm. 
    ($\mathcal{S}_p$: Pial SDF, $\mathcal{S}_w$: White matter SDF, $\mathcal{S}_c$: Cortex SDF, $\mathcal{S}_d$: Joint cortex DF, $\mathcal{E}$: Edge map, $\mathcal{R}$: Cortical ribbon mask)}
    \begin{threeparttable}
        
    \begin{tabular}{llcccccc}
    \toprule
    && \multicolumn{4}{c}{ASSD$\downarrow$} \\
    \cmidrule(lr){3-6}
    && \multicolumn{2}{c}{Left Hemisphere} & \multicolumn{2}{c}{Right Hemisphere} \\
    \cmidrule(lr){3-4}
    \cmidrule(lr){5-6}
    Source & Aux. Condition
    & White & Pial & White & Pial \\
    \midrule

    $\mathcal{S}_p$ & -- 
    & 0.375{\tiny\textpm0.066}
    & \underline{0.233{\tiny\textpm0.021}}
    & 0.386{\tiny\textpm0.068}
    & \underline{0.236{\tiny\textpm0.022}}
    \\

    $\mathcal{S}_w$ & -- 
    & \underline{0.289{\tiny\textpm0.066}}
    & 0.375{\tiny\textpm0.031}
    & 0.303{\tiny\textpm0.067}
    & 0.376{\tiny\textpm0.034}
    \\
    
    $\mathcal{E}$ & -- 
    & 0.351{\tiny\textpm0.037}
    & 0.307{\tiny\textpm0.023}
    & 0.369{\tiny\textpm0.042}
    & 0.307{\tiny\textpm0.024}
    \\

    $\mathcal{R}$ & -- 
    & 0.336{\tiny\textpm0.034}
    & 0.279{\tiny\textpm0.021}
    & 0.335{\tiny\textpm0.037}
    & 0.280{\tiny\textpm0.022}
    \\

    $\mathcal{S}_d$ & -- 
    & 0.333{\tiny\textpm0.039}
    & 0.250{\tiny\textpm0.022}
    & 0.342{\tiny\textpm0.040}
    & 0.249{\tiny\textpm0.021}
    \\

    $\mathcal{S}_p \cap \mathcal{S}_w$ & -- 
    & 0.361{\tiny\textpm0.087}
    & 0.365{\tiny\textpm0.033}
    & 0.373{\tiny\textpm0.088}
    & 0.366{\tiny\textpm0.033}
    \\
  
    $\mathcal{S}_p \cup \mathcal{S}_w$ & -- 
    & 0.416{\tiny\textpm0.083}
    & 0.242{\tiny\textpm0.025}
    & 0.429{\tiny\textpm0.087}
    & 0.244{\tiny\textpm0.026}
    \\

    $\mathcal{S}_p \mathbin{\sqcup_{10}} \mathcal{S}_w$ & -- 
    & 0.316{\tiny\textpm0.076}
    & 0.261{\tiny\textpm0.025}
    & 0.331{\tiny\textpm0.082}
    & 0.262{\tiny\textpm0.027}
    \\

    $\mathcal{S}_p \mathbin{\sqcup_{100}} \mathcal{S}_w$ & -- 
    & 0.332{\tiny\textpm0.065}
    & 0.256{\tiny\textpm0.024}
    & 0.339{\tiny\textpm0.072}
    & 0.256{\tiny\textpm0.025}
    \\

    $\mathcal{S}_p, \mathcal{S}_w$\tnote{*} & -- 
    & 0.394{\tiny\textpm0.074}
    & 0.535{\tiny\textpm0.064}
    & 0.418{\tiny\textpm0.079}
    & 0.553{\tiny\textpm0.071}
    \\

    $\mathcal{S}_c$ & -- 
    & 0.381{\tiny\textpm0.066}
    & 0.245{\tiny\textpm0.026}
    & 0.390{\tiny\textpm0.070}
    & 0.247{\tiny\textpm0.026}
    \\

    \midrule

    $\mathcal{R}$ & $\mathcal{S}_p, \mathcal{S}_w$
    & 0.293{\tiny\textpm0.027}
    & 0.262{\tiny\textpm0.018}
    & 0.305{\tiny\textpm0.031}
    & 0.261{\tiny\textpm0.016}
    \\

    $\mathcal{R}$  & $\mathcal{S}_p, \mathcal{S}_w, \mathcal{E}$
    & 0.301{\tiny\textpm0.038}
    & 0.263{\tiny\textpm0.021}
    & 0.305{\tiny\textpm0.040}
    & 0.264{\tiny\textpm0.022}
    \\
    \midrule

    $\mathcal{S}_d$  & $\mathcal{E}$
    & 0.338{\tiny\textpm0.034}
    & 0.305{\tiny\textpm0.022}
    & 0.347{\tiny\textpm0.037}
    & 0.310{\tiny\textpm0.023}
    \\

    $\mathcal{S}_d$  & $\mathcal{R}$
    & 0.305{\tiny\textpm0.037}
    & 0.268{\tiny\textpm0.034}
    & 0.316{\tiny\textpm0.040}
    & 0.267{\tiny\textpm0.038}
    \\

    $\mathcal{S}_d$  & $\mathcal{S}_p, \mathcal{S}_w$
    & 0.310{\tiny\textpm0.058}
    & \textbf{0.231{\tiny\textpm0.015}}
    & 0.326{\tiny\textpm0.063}
    & \textbf{0.234{\tiny\textpm0.015}}
    \\

    \midrule

    $\mathcal{S}_c$  & $\mathcal{E}$
    & 0.326{\tiny\textpm0.036}
    & 0.241{\tiny\textpm0.021}
    & 0.339{\tiny\textpm0.039}
    & 0.243{\tiny\textpm0.022}
    \\

    $\mathcal{S}_c$ & $\mathcal{R}$
    & 0.305{\tiny\textpm0.030}
    & 0.261{\tiny\textpm0.021}
    & 0.312{\tiny\textpm0.033}
    & 0.262{\tiny\textpm0.020}
    \\

    $\mathcal{S}_c$ & $\mathcal{S}_p, \mathcal{S}_w$
    & 0.310{\tiny\textpm0.060}
    & 0.236{\tiny\textpm0.020}
    & 0.321{\tiny\textpm0.064}
    & 0.239{\tiny\textpm0.022}
    \\

    $\mathcal{S}_c$ & $\mathcal{E}, \mathcal{R}$
    & \underline{0.289{\tiny\textpm0.031}}
    & 0.252{\tiny\textpm0.016}
    & \underline{0.294{\tiny\textpm0.034}}
    & 0.254{\tiny\textpm0.016}
    \\

    $\mathcal{S}_c$ & $\mathcal{S}_p, \mathcal{S}_w, \mathcal{E}$
    & 0.294{\tiny\textpm0.043}
    & 0.236{\tiny\textpm0.017}
    & 0.311{\tiny\textpm0.047}
    & 0.238{\tiny\textpm0.018}
    \\

    $\mathcal{S}_c$ & $\mathcal{S}_p, \mathcal{S}_w, \mathcal{R}$
    & 0.305{\tiny\textpm0.036}
    & 0.255{\tiny\textpm0.019}
    & 0.308{\tiny\textpm0.037}
    & 0.255{\tiny\textpm0.019}
    \\

    $\mathcal{S}_c$ & $\mathcal{S}_p, \mathcal{S}_w, \mathcal{E}, \mathcal{R}$
    & \textbf{0.282{\tiny\textpm0.032}}
    & 0.250{\tiny\textpm0.017}
    & \textbf{0.287{\tiny\textpm0.035}}
    & 0.250{\tiny\textpm0.017}
    \\

    \bottomrule
    
    \end{tabular}
    \begin{tablenotes}\scriptsize
    \item[*] This model requires two parallel Brownian bridge processes.    
    \end{tablenotes}
    \end{threeparttable}
    
    \label{tab:ablation}
\end{table}

\subsection{Analysis of shape-conditioning strategies} \label{sec:ablation-results}

In this section, we examine the impact of shape conditioning on the geometric consistency in Cor2Vox, focusing on the choice of shape representations in the source shape domain and the auxiliary shape conditions. See \Cref{sec:cortex-to-image-model} for methodological details.

\subsubsection{Analysis of the primary shape condition}

As summarized in~\Cref{tab:ablation}, we first assessed each individual shape condition: pial surface signed distance function (SDF) $\mathcal{S}_p$, white matter surface SDF $\mathcal{S}_w$, edge map $\mathcal{E}$, cortical ribbon mask $\mathcal{R}$, and cortex SDF $\mathcal{S}_c$, as single-source domains in the Brownian bridge diffusion process. 
Conditioning solely on the Pial SDF $\mathcal{S}_p$ yielded the best pial surface accuracy (0.233\textpm0.021 mm ASSD for the left hemisphere), and similarly, white matter SDF $\mathcal{S}_w$ yielded the best white matter accuracy (0.289\textpm0.066 mm). However, in both cases, the model performed poorly on the respective other surface, which was not explicitly specified.
The cortical ribbon mask ($\mathcal{R}$) provided high accuracy on the white matter generation while falling short on the pial surface synthesis. 

Next, we investigated different strategies for combining the two complementary surface SDFs, pial surface $\mathcal{S}_p$ and white matter surface $\mathcal{S}_w$. 
Specifically, we evaluated intersection ($\mathcal{S}_p \cap \mathcal{S}_w$), union ($\mathcal{S}_p \cup \mathcal{S}_w$), soft union ($\mathcal{S}_p \mathbin{\sqcup_k} \mathcal{S}_w$, parameterized by a sharpness factor $k$), and joint distance function (DF) $\mathcal{S}_d$, which is formed by summing the distance functions of both surfaces. These combination strategies resulted in poorer performance on at least one surface compared to single-surface conditioning, except $\mathcal{S}_d$ and $\mathcal{S}_p \mathbin{\sqcup_{100}} \mathcal{S}_w$ reaching on-par performance.
Furthermore, two parallel Brownian bridge processes, conditioned on both $\mathcal{S}_p$ and $\mathcal{S}_w$, resulted in the worst pial surface accuracy (0.535\textpm0.064 mm), indicating that two simultaneous, uncoordinated Brownian bridge processes introduce inaccuracies.

\subsubsection{Analysis of auxiliary shape conditions}

Finally, we assessed the impact of providing various shape conditions as auxiliary conditions in Cor2Vox. Those auxiliary conditions are provided as concatenated inputs to the generative model to further guide the generation based on the primary condition, cf.\,\Cref{fig:architecture}. 
Using the cortex SDF ($\mathcal{S}_c$) as the primary condition, we progressively introduced combinations of auxiliary inputs ($\mathcal{S}_p$, $\mathcal{S}_w$, $\mathcal{E}$, $\mathcal{R}$).
Incorporating more shape cues consistently improved geometric accuracy, as reported in \Cref{tab:ablation}, with the combination of all four conditions ($\mathcal{S}_p$, $\mathcal{S}_w$, $\mathcal{E}$, $\mathcal{R}$) yielding the best overall consistency. In particular, it enhanced white matter accuracy while maintaining comparable pial surface performance, consistently across hemispheres.
We also evaluated using either $\mathcal{R}$ or $\mathcal{S}_d$ as the primary condition in combination with auxiliary inputs, both of which outperformed their original scores (without auxiliary conditioning). In summary, these results confirm the effectiveness of adopting $\mathcal{S}_c$ as the primary condition in Cor2Vox, and they highlight the benefit of incorporating complementary shape information via the proposed auxiliary conditioning.

\section{Discussion}

Our study revealed flaws in conventional DDPMs~\cite{ddpm}, arguably the de facto standard architecture for high-dimensional image generation, when applied to unconditional 3D brain image generation, cf.\ \Cref{sec:plausibility-results}. Specifically, we found that 3D-DDPMs do not preserve characteristic grooves in the cerebral cortex, such as the central and superior temporal sulci. These irregularities break standardized atlas-based brain parcellation, which is crucial for many downstream neuroscience applications~\cite{Lawrence2021parcellation}.
Our findings align with prior work advocating for rigorous evaluation of synthetic brain MRIs~\cite{Wu2024evaluating}, extending these efforts beyond volume-based metrics such as Cohen’s d to include advanced cortical surface reconstruction. In Cor2Vox, we addressed these issues by incorporating an explicit shape model of the cerebral cortex. This guarantees anatomical plausibility and, hence, compatibility with surface-based analysis pipelines and standardized parcellation schemes.

When it comes to brain morphology in synthetic MRIs, conditioning on low-dimensional demographic and clinical covariates, such as age, sex, and diagnosis, is a popular approach~\cite{Peng2024metadataconditioned,Tudosiu2024morphology-preserving}. However, the generative capacity of these methods remains, by design, limited to implicit associations between the covariates and structural observations that were included in the training. In addition, such models are prone to inheriting existing biases as their outputs reflect statistical correlations in the training data~\cite{Kocak2024biasinAI}. In contrast, Cor2Vox readily decouples from the training data, as evidenced by its ability to synthesize images for frontotemporal dementia, a condition not contained in the ADNI training set.
Moreover, our results on cortical atrophy simulation showed that Cor2Vox accurately reflected simulated changes imposed on cortical surfaces, cf.\ \Cref{sec:atrophy-results}. This capability allows researchers to systematically generate edge cases, explore hypothetical or extreme conditions, and stress-test models under controlled variations, an opportunity usually not provided by existing conditional generative models. The reported mean absolute error between introduced and recovered cortical thickness changes of 0.14~mm is in the range of the expected variability of morphometric neuroimaging pipelines for cortical thickness estimation~\cite{Han2006reliability,Rusak2022benchmark}.

The capability of capturing continuous simulations also distinguishes Cor-2Vox from existing shape-based generative models, such as Pix2Pix~\cite{pix2pix2017}, BBDM~\cite{li_bbdm_2023}, and Med-DDPM~\cite{Dorjsembe2024medddpm}. Nevertheless, these approaches represent important baselines for our approach. In our experiments, Cor2Vox outperformed all of them significantly in terms of geometric consistency with the cortical shape priors at comparable image fidelity metrics, cf.\ \Cref{sec:comparison}. Our ablation study suggests that the combination of the cortex SDF-based Brownian bridge process together with the auxiliary cortex representations (pial SDF, white matter SDF, cortex edge map, and cortical ribbon mask) is the key driver for the high accuracy in Cor2Vox, cf.\ \Cref{sec:ablation-results}.
Importantly, all of our generated images passed the recommended quality threshold of SynthSeg+~\cite{billot2023synthseg+}, cf.\ \Cref{fig:generation_quality}, indicating that the synthesized images meet established standards for segmentation reliability, in all structures, not only in cortical gray matter, and are suitable for downstream neuroimaging analyses.

While our framework demonstrated strong performance in generating anatomically realistic brain MRIs conditioned on complex cortical surfaces, it is currently limited to this imaging modality. Specifically, we have not yet explored conditioning on additional anatomical or pathological structures beyond the cerebral cortex, such as subcortical regions or tumors. 
Nonetheless, we believe the modularity of our framework lends itself well to extension, and conditioning on simpler shapes, such as those found in subcortical structures or lesion masks, should be feasible with dedicated architectural changes. Future work will investigate these directions to broaden the applicability of shape-grounded generative modeling.

\section{Conclusion}

In summary, Cor2Vox advanced 3D medical image generation through surface-based modeling of anatomical shape. By incorporating a population-scale, mesh-based shape model, Cor2Vox enabled anatomically faithful synthesis of 3D brain MRIs with high geometric fidelity using fine-grained cortical surfaces as structural priors. Our experiments showed strict consistency between the synthesized images and the conditional inputs, demonstrating the capability to perform sub-voxel simulations of cortical atrophy and effectively disentangle brain shape and appearance.
Cor2Vox's ability to produce a variety of phenotypes via anatomically grounded sampling and simulation, without retraining, highlighted its potential as a powerful tool for data augmentation, disease progression modeling, and cross-site harmonization in neuroimaging studies. Together, these results demonstrated that explicit surface-based shape modeling provides a principled and effective foundation for controllable, biologically plausible 3D medical image generation, thereby addressing key limitations in previous generative models.

\section*{Data availability}

Data from ADNI are available from the LONI website at \url{https://ida.loni.usc.edu} upon registration and compliance with the data usage agreement. 
UK Biobank (UKB) data can be requested via the UKB website (\url{https://www.ukbiobank.ac.uk/}) following their application process. The in-house dataset can be shared upon request.
We used the Montreal Neuroimaging Institute MNI152 template for image processing purposes; it is available for download at \url{http://www.bic.mni.mcgill.ca/ ServicesAtlases/ICBM152NLin2009}.

\section*{Code availability}

Our code is available on GitHub (\url{https://github.com/ai-med/Cor2Vox}).

\section*{Acknowledgments} 

Y.L. was supported by the Munich Center for Machine Learning (MCML). C.W. received funding from the German Research Foundation (DFG) and the DAAD programme Konrad Zuse Schools of Excellence in Artificial Intelligence, sponsored by the Federal Ministry of Research, Technology, and Space. 

Data collection and sharing for this project was funded by the Alzheimer’s Disease Neuroimaging Initiative (ADNI) (National Institutes of Health Grant U01 AG024904) and DOD ADNI (Department of Defense award number W81XWH-12-2-0012). ADNI is funded by the National Institute on Aging, the National Institute of Biomedical Imaging and Bioengineering, and through generous contributions from the following: Alzheimer’s Association; Alzheimer’s Drug Discovery Foundation; Araclon Biotech; BioClinica, Inc.; Biogen Idec Inc.; BristolMyers Squibb Company; Eisai Inc.; Elan Pharmaceuticals, Inc.; Eli Lilly and Company; EuroImmun; F. Hoffmann-La Roche Ltd and its affiliated company Genentech, Inc.; Fujirebio; GE Healthcare; ; IXICO Ltd.; Janssen Alzheimer Immunotherapy Research \& Development, LLC.; Johnson \& Johnson Pharmaceutical Research \& Development LLC.; Medpace, Inc.; Merck \& Co., Inc.; Meso Scale Diagnostics, LLC.; NeuroRx Research; Neurotrack Technologies; Novartis Pharmaceuticals Corporation; Pfizer Inc.; Piramal Imaging; Servier; Synarc Inc.; and Takeda Pharmaceutical Company. The Canadian Institutes of Health Research is providing funds to support ADNI clinical sites in Canada. Private sector contributions are facilitated by the Foundation for the National Institutes of Health (\url{www.fnih.org}). The grantee organization is the Northern California Institute for Research and Education, and the study is coordinated by the Alzheimer’s Disease Cooperative Study at the University of California, San Diego. ADNI data are disseminated by the Laboratory for Neuro Imaging at the University of Southern California.

Data used in the preparation of this article were obtained from the UK Biobank
Resource under Application No.\ 34479.

We thank Dr.~med.~Dennis Hedderich for providing clinical MRI scans from 10 patients, which supported the evaluation of Cor2Vox on frontotemporal dementia.

We further gratefully acknowledge the scientific support and resources of the AI service infrastructure LRZ AI Systems provided by the Leibniz Supercomputing Centre (LRZ).

\clearpage

\bibliography{bibliography}

\clearpage

\appendix

\section*{Supplementary Material}

\renewcommand{\thefigure}{S\arabic{figure}}
\setcounter{figure}{0}
\renewcommand{\thetable}{S\arabic{table}}
\setcounter{table}{0}

\begin{figure}[hp]
    \centering
    \includegraphics[width=0.95\linewidth]{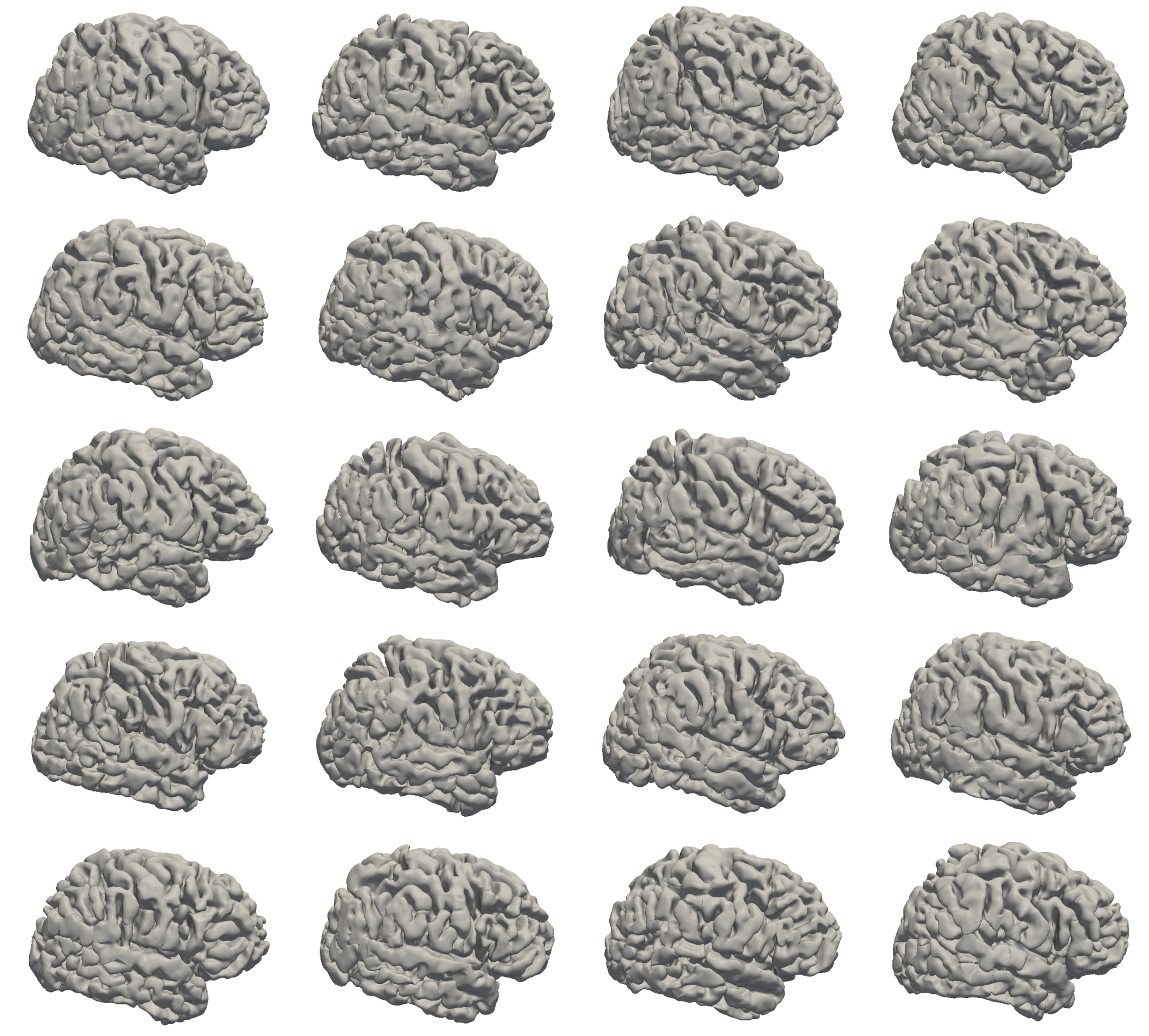}
    \caption{Reconstructed cortical surfaces of the right hemisphere from 20 arbitrarily sampled brain MRIs using 3D-DDPM. They suffer from flaws in anatomical plausibility, such as irregular, discontinuous grooves and misplaced folds.}
    \label{fig:ddpm-samples}
\end{figure}

\clearpage

\begin{figure}
    \centering
    \includegraphics[width=0.98\linewidth]{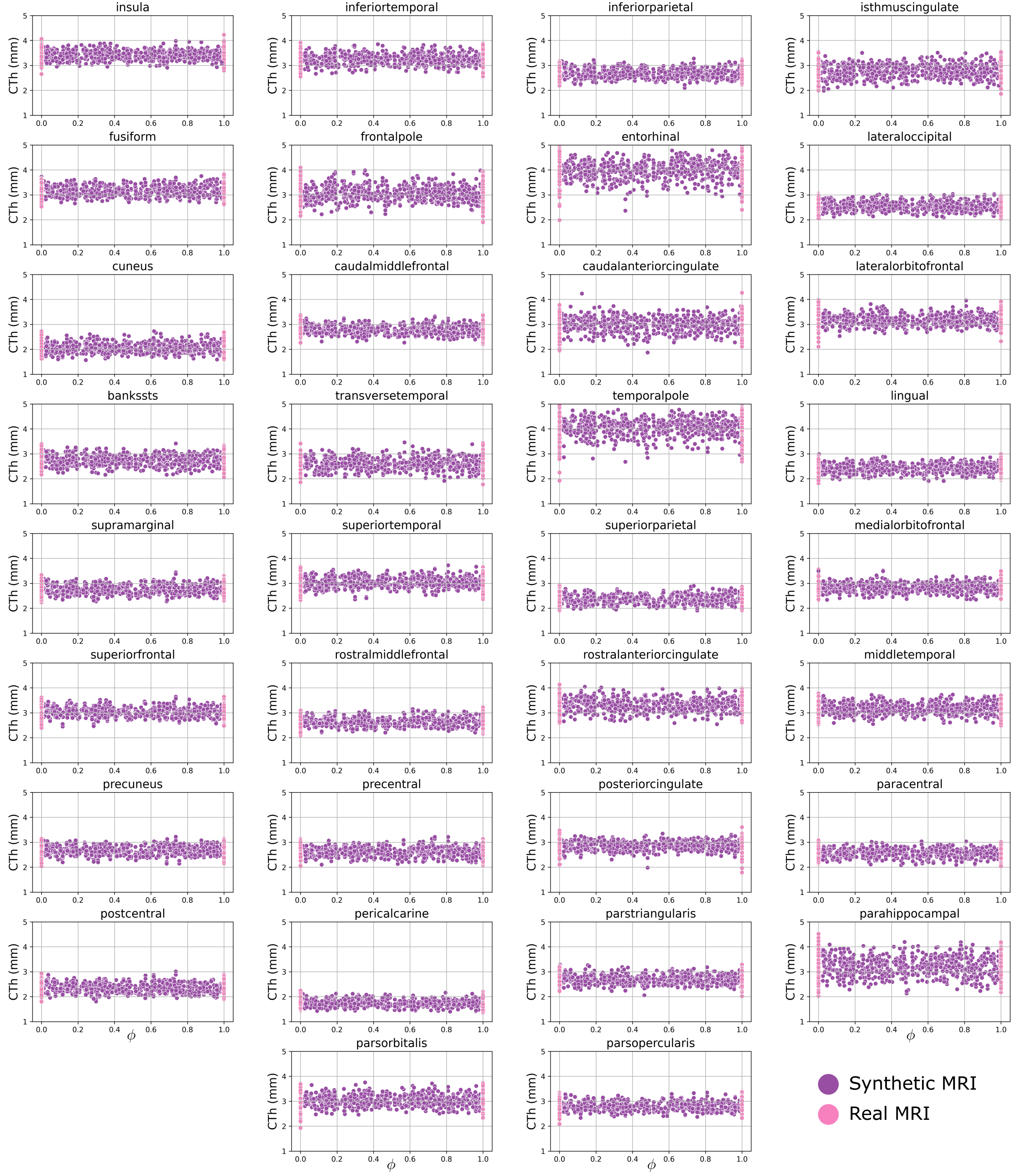}
    \caption{Comparison of regional cortical thickness (CTh) in real and synthetic MRIs by Cor2Vox based on the Desikan-Killany atlas. Cor2Vox preserves the magnitude and spread of cortical thickness observed in real-world data across all 34 DK regions.}
    \label{fig:local-cth-all}
\end{figure}

\clearpage

\begin{figure}
\includegraphics[width=0.98\linewidth]{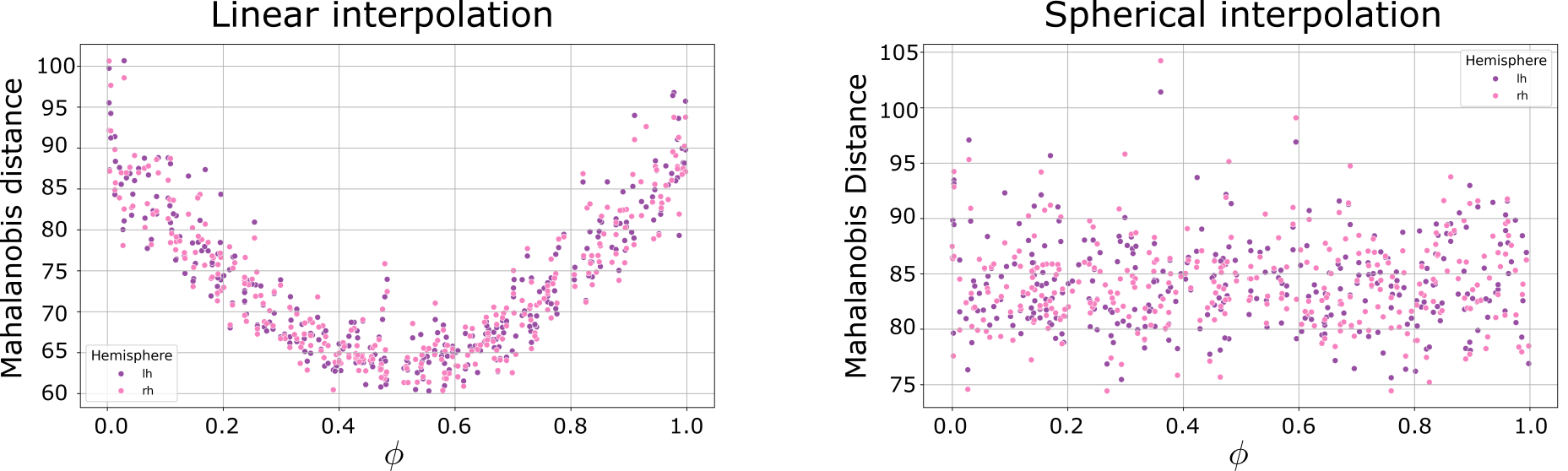}
\caption{Mahalanobis distance of cortical shapes to the population mean in PCA latent space for linear (left) and spherical (right) interpolation. $0<\phi<1$ denotes the interpolation factor, with $\phi=0/1$ corresponding to real samples. While linear interpolation tends to drift toward the mean, yielding implausibly smooth shapes, spherical interpolation circumvents this issue, producing more realistic cortex shapes.}
\label{fig:mahalanobis}
\end{figure}

\clearpage

\begin{table}[t]
    \setlength{\tabcolsep}{1.2pt}
    \renewcommand\bfdefault{b}
    \centering
    \caption{Image fidelity comparison of implemented methods for 3D brain MRI generation. Structural similarity index measure (SSIM$\uparrow$) and peak signal-to-noise ratio (PSNR$\uparrow$) quantify whole-brain image fidelity, and average symmetric surface distance (ASSD$\downarrow$, mm) measures geometric consistency of white matter (WM) and pial cortical surfaces. Values represent the mean and standard deviation across all samples in our test set ($n=323$).}
    \begin{threeparttable}
    \begin{tabular}{lcccccc}
    \toprule
    &&& \multicolumn{4}{c}{ASSD$\downarrow$} \\
    \cmidrule(lr){4-7}
    & \multicolumn{2}{c}{Whole Brain} & \multicolumn{2}{c}{Left Hemisphere} & \multicolumn{2}{c}{Right Hemisphere} \\
    \cmidrule(lr){2-3}
    \cmidrule(lr){4-5}
    \cmidrule(lr){6-7}
    Model 
    & SSIM$\uparrow$ & PSNR$\uparrow$ & WM & Pial & WM & Pial \\

    \midrule
    
    Pix2Pix$_\mathcal{E}$\cite{pix2pix2017}\tnote{*}
    & 0.891{\tiny\textpm0.011}
    & \textbf{24.76{\tiny\textpm1.96}}
    & 0.784{\tiny\textpm0.190}
    & 0.853{\tiny\textpm0.227}
    & 0.760{\tiny\textpm0.174}
    & 0.804{\tiny\textpm0.195}
    \\

    Med-DDPM$_\mathcal{E}$\cite{Dorjsembe2024medddpm}
    & 0.894{\tiny\textpm0.015}
    & 22.38{\tiny\textpm2.63}
    & 0.506{\tiny\textpm0.042}
    & 0.437{\tiny\textpm0.027}
    & 0.516{\tiny\textpm0.043}
    & 0.436{\tiny\textpm0.026}
    \\

    BBDM$_\mathcal{E}$\cite{li_bbdm_2023}\tnote{*}
    & 0.898{\tiny\textpm0.020}
    & 19.95{\tiny\textpm2.61}
    & 0.349{\tiny\textpm0.032}
    & 0.307{\tiny\textpm0.021}
    & 0.368{\tiny\textpm0.035}
    & 0.307{\tiny\textpm0.021}
    \\
    
    \midrule

    Pix2Pix$_\mathcal{R}$\cite{pix2pix2017}\tnote{*}
    & 0.885{\tiny\textpm0.013}
    & 23.69{\tiny\textpm2.06}
    & 0.697{\tiny\textpm0.137}
    & 0.691{\tiny\textpm0.155}
    & 0.698{\tiny\textpm0.183}
    & 0.680{\tiny\textpm0.221}
    \\

    Med-DDPM$_\mathcal{R}$\cite{Dorjsembe2024medddpm}
    & 0.899{\tiny\textpm0.016}
    & 22.30{\tiny\textpm2.80}
    & 0.359{\tiny\textpm0.043}
    & 0.347{\tiny\textpm0.034}
    & 0.373{\tiny\textpm0.106}
    & 0.352{\tiny\textpm0.101}
    \\

    BBDM$_\mathcal{R}$\cite{li_bbdm_2023}\tnote{*}
    & 0.898{\tiny\textpm0.021}
    & 19.67{\tiny\textpm2.54}
    & 0.328{\tiny\textpm0.037}
    & 0.283{\tiny\textpm0.034}
    & 0.338{\tiny\textpm0.128}
    & 0.287{\tiny\textpm0.121}
    \\

    \midrule

    Cor2Vox{\scriptsize/DDPM}
    & 0.902{\tiny\textpm0.015}
    & 23.13{\tiny\textpm2.71}
    & 0.335{\tiny\textpm0.038}
    & 0.300{\tiny\textpm0.033}
    & 0.348{\tiny\textpm0.048}
    & 0.305{\tiny\textpm0.037}
    \\
    
    Cor2Vox
    & \textbf{0.906{\tiny\textpm0.018}}
    & 21.10{\tiny\textpm2.74}
    & \textbf{0.283{\tiny\textpm0.029}}
    & \textbf{0.251{\tiny\textpm0.019}}
    & \textbf{0.289{\tiny\textpm0.031}}
    & \textbf{0.251{\tiny\textpm0.017}}
    \\

    \bottomrule
    
    \end{tabular}

    \begin{tablenotes} \scriptsize
            \item[*] Method adapted for 3D image generation.
        \end{tablenotes}
    \end{threeparttable}
    \label{tab:qual-acc}
\end{table}

\clearpage

\begin{figure}[hp]
    \centering
    \includegraphics[width=0.95\linewidth]{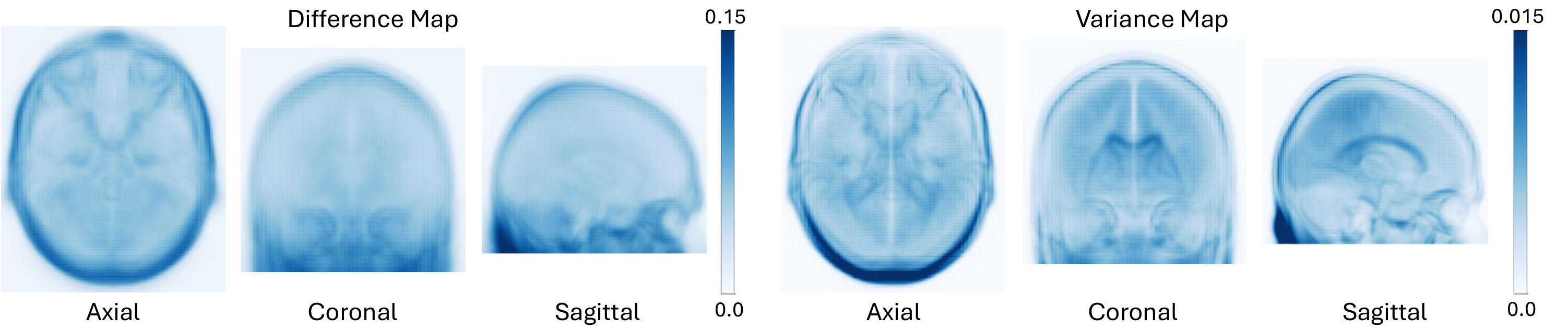}
    \caption{Generation variability in Cor2Vox outputs. We present the mean absolute voxel-wise difference between synthetic MRIs generated from Cor2Vox and corresponding original MRIs (with the same cortical shape), as well as voxel-wise variance in synthetic MRIs across five random seeds, based on the ADNI test set ($n=323$). Values are averaged along the three anatomical axes. 
    Darker colors indicate higher difference/variability.
    }
    \label{fig:varibaility}
\end{figure}

\clearpage

\begin{figure}[hp]
    \centering
    \includegraphics[width=0.9\linewidth]{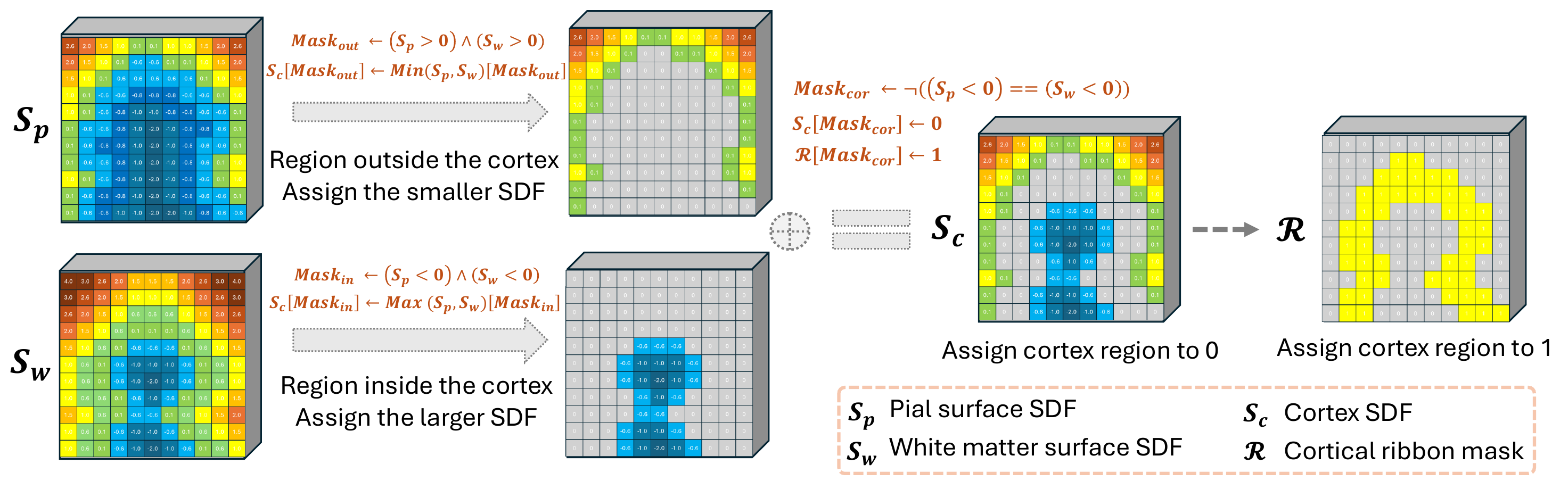}
    \caption{Generation of cortex SDF ($\mathcal{S}_c$) and cortical ribbon mask ($\mathcal{R})$.}
    \label{fig:cortexsdf}
\end{figure}

\end{document}